\documentclass[10pt,twocolumn,letterpaper]{article}

\usepackage[pagenumbers]{cvpr} 

\definecolor{cvprblue}{rgb}{0.21,0.49,0.74}
\usepackage[pagebackref,breaklinks,colorlinks,allcolors=cvprblue]{hyperref}

\usepackage[accsupp]{axessibility}
\usepackage{amsthm}

\newcommand{\minisection}[1]{\vspace{0.025in}\noindent{\bf #1}\hspace{0.10em}}
\usepackage{multirow}
\usepackage{bbding}
\usepackage{mathtools}
\usepackage[table]{xcolor}
\colorlet{myhighlight}{green!25}
\definecolor{LightPurple}{rgb}{0.88,0.88,1}
\definecolor{celadon}{rgb}{0.75, 0.88, 0.85}
\newcommand{\hlrow}{\rowcolor{celadon}}

\usepackage{algorithm} 
\usepackage{algpseudocode} 
\usepackage{amsmath} 
\usepackage{amssymb} 
\usepackage{wrapfig}
\usepackage{tocloft}
\usepackage{etoolbox}

\def\paperID{} 
\def\confName{CVPR}
\def\confYear{2026}

\title{IsoCLIP: Decomposing CLIP Projectors for Efficient Intra-modal Alignment}

\author{
Simone Magistri$^{1}$\thanks{Corresponding author.} \and
Dipam Goswami$^{2,3}$ \and
Marco Mistretta$^{1}$ \and 
Bartłomiej Twardowski$^{2,3,4}$ \and 
Joost van de Weijer$^{2,3}$ \quad Andrew D. Bagdanov$^{1}$
\\
$^{1}$ Media Integration and Communication Center (MICC), University of Florence, Italy \\
$^{2}$Department of Computer Science, Universitat Autònoma de Barcelona, Spain \\ 
$^{3}$Computer Vision Center, Barcelona, Spain \quad 
$^{4}$IDEAS Research Institute, Warsaw, Poland \\
{\tt\small \{simone.magistri, marco.mistretta, andrew.bagdanov\}@unifi.it} \\
{\tt\small \{dgoswami, btwardowski, joost\}@cvc.uab.cat}
}

\begin{document}
\maketitle

\begin{abstract}
Vision-Language Models like CLIP are extensively used for inter-modal tasks which involve both visual and text modalities. However, when the individual modality encoders are applied to inherently intra-modal tasks like image-to-image retrieval, their performance suffers from the intra-modal misalignment. In this paper we study intra-modal misalignment in CLIP with a focus on the role of the projectors that map pre-projection image and text embeddings into the shared embedding space. By analyzing the form of the cosine similarity applied to projected features, and its interaction with the contrastive CLIP loss, we show that there is an inter-modal operator responsible for aligning the two modalities during training, and a second, intra-modal operator that only enforces intra-modal normalization but does nothing to promote intra-modal alignment. Via spectral analysis of the inter-modal operator, we identify an approximately isotropic subspace in which the two modalities are well-aligned, as well as anisotropic directions specific to each modality. We demonstrate that this aligned subspace can be directly obtained from the projector weights and that removing the anisotropic directions improves intra-modal alignment. Our experiments on intra-modal retrieval and classification benchmarks show that our training-free method reduces intra-modal misalignment, greatly lowers latency, and outperforms existing approaches across multiple pre-trained CLIP-like models. The code is publicly available at: \small{\href{https://github.com/simomagi/IsoCLIP}{\mbox{\url{https://github.com/simomagi/IsoCLIP}}}}.
\end{abstract}

\section{Introduction}
\label{sec:intro}

\begin{figure*}
\centering
\includegraphics[width=0.98\linewidth]{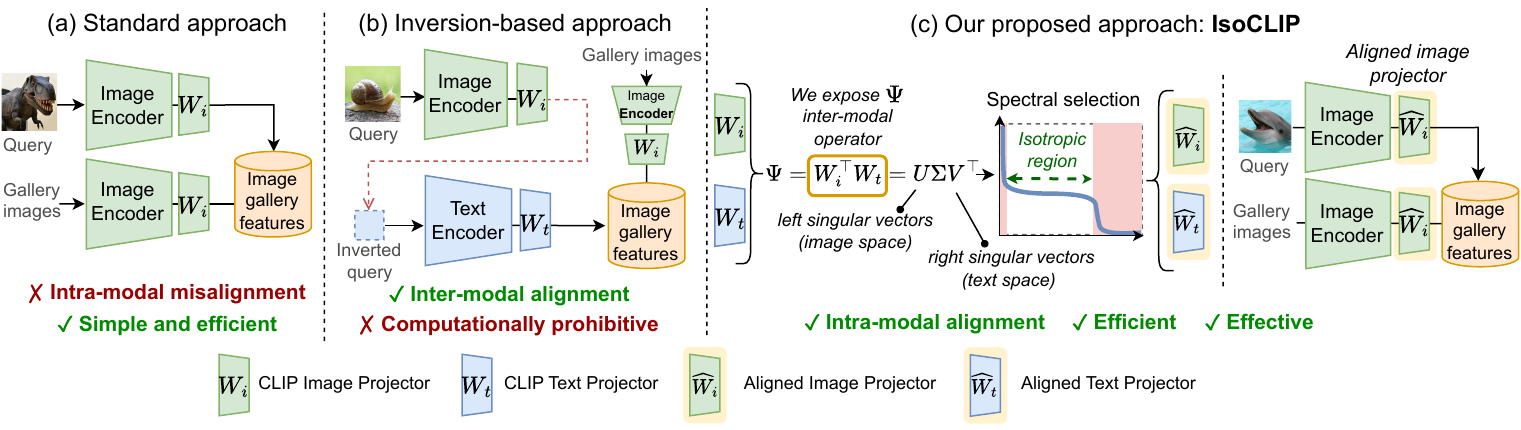}
\caption{\textbf{Overview of intra-modal retrieval with CLIP.} (a) The standard approach simply compares the cosine similarities computed after applying projector $W_i$ to query and gallery image embeddings, which is sub-optimal due to \textit{intra-modal misalignment}. (b) To circumvent misalignment, inversion approaches~\cite{mistretta2025cross} convert the query image embeddings to text embeddings by iteratively optimizing pseudo-tokens -- an expensive operation that incurs high latency -- and then computes inter-modal cosine similarities for retrieval. (c) We identify an inter-modal operator $\Psi=W_i^{\top} W_t$ fundamental to CLIP cosine similarity computations. We propose IsoCLIP, which uses only an isotropic region of the spectrum of $\Psi$ to align the projector weights along well-aligned directions between modalities. Then these \textit{aligned} projectors are used to map the query and gallery embeddings. IsoCLIP exploits the properties of the CLIP projectors and does not add any latency to process while yielding more optimal intra-modal cosine similarities and significantly improved intra-modal performance.}
\vspace{-10pt}
\label{fig:teaser}
\end{figure*}

Pre-trained Vision-Language Models (VLMs) like CLIP~\cite{radford2021learning} are widely used for applications ranging from image and text retrieval~\cite{baldrati2022effective,sain2023clip}, to open-vocabulary
segmentation~\cite{gui2024knnclip,yu2023convolutions}, generalized category discovery ~\cite{Wang_2025_CVPR,caselli2026spectralgcd} and visual question answering~\cite{song2022clip,parelli2023clip}. CLIP is trained contrastively on a massive dataset of image-text pairs to align image and text representations from the vision and text encoders by projecting them in a shared embedding space. This joint embedding space has fostered a broad range of inter-modal applications leveraging the power of the semantically-aligned image and text encoders.

While primarily designed for inter-modal tasks like zero-shot object recognition, CLIP has been applied to intra-modal tasks due to its rich pre-trained image and text encoders. Several works use the CLIP image encoder for image-to-image retrieval~\cite{mistretta2025cross,kordopatis2025ilias,nara2024revisiting,nakata2022revisiting,schall2024optimizing}, sketch-based image retrieval~\cite{sain2023clip}, and the CLIP text encoder for text-to-text retrieval~\cite{xiao2024jina,mistretta2025cross}. Furthermore, CLIP image encoders have also been extensively used for computing intra-modal image-to-image similarities for tasks like image classification~\cite{zhang2022tip,geirhos2024towards,udandarao2023sus,wei2022mvp} and text-to-image generation~\cite{ruiz2023dreambooth,gal2022image}. While it is desirable to have a multi-modal model which can also be used for intra-modal tasks when required, the \textit{intra-modal misalignment} inherent in CLIP -- as pointed out in~\cite{mistretta2025cross,udandarao2023sus,yi2024leveraging} -- results in sub-optimal performance.

Intra-modal misalignment manifests because the contrastive training loss of CLIP maximizes only inter-modal cosine similarities while ignoring the intra-modal ones. \citet{mistretta2025cross} showed that approaching intra-modal tasks inter-modally can mitigate intra-modal misalignment. To demonstrate this, they introduced two \textit{modality inversion} techniques -- Optimization-based Text Inversion (OTI) and Optimization-based Visual Inversion (OVI)  -- which map a feature into the complementary modality.

In image-to-image retrieval, OTI converts a query image feature into a query text feature, which is then compared with the original gallery image features. Conversely, in text-to-text retrieval, OVI performs the opposite mapping, inverting a query text feature into an image feature. Although they mitigate intra-modal misalignment, these techniques have significant limitations that hinder practical applicability. In particular, they require very many optimization steps and forward passes through the target modality encoder, and their performance is strongly dependent on the number of optimization steps which are difficult to determine a priori.

In this paper, we analyze the training loss of CLIP and its interaction with the projection heads used to project pre-projection feature embeddings into the aligned space. We show that there is an \textit{inter-modal operator} hidden in the cosine similarity which connects the two pre-projection spaces and is responsible for aligning the image and text embeddings. We additionally show that, during backpropagation of the  contrastive CLIP loss, an \textit{intra-modal operator} is unveiled that guarantees normalization, but does nothing to promote intra-modal alignment. Via Singular Value Decomposition of the inter-modal operator, we identify an approximately isotropic subspace in which the two modalities are well-aligned and show how the two modalities behave in the different regions of the singular value spectra. Finally, we show (see~\cref{fig:teaser}) that restricting projectors to these identified isotropic directions yields more discriminative cosine similarities and significant improvement on intra-modal tasks like image-to-image retrieval, text-to-text retrieval. To summarize, our main contributions are:
\begin{itemize}
    \item we analyze the role of CLIP projectors and dissect their interactions with the cosine similarity and contrastive loss, which reveals an inter-modal operator responsible for aligning modalities and an intra-modal operator responsible only for normalization;
    \item via spectral analysis of the inter-modal operator, we identify an approximately isotropic subspace in the middle band of the spectrum in which the modalities are well-aligned and anisotropic bands at the top and bottom of the spectrum specific to each modality;
    \item we propose IsoCLIP, which enhances intra-modal alignment by retaining only well-aligned directions between text and images in the inter-modal operator while discarding anisotropic, modality-specific directions; and 
    \item we experimentally demonstrate that IsoCLIP greatly reduces latency and consistently improves performance on a range of intra-modal tasks, including image-to-image retrieval, text-to-text retrieval across multiple datasets.
\end{itemize}

\section{Related Work}
In this section we review work from the recent literature most related to our contributions.

\minisection{Vision-language models.} Contrastively trained VLMs~\cite{radford2021learning,zhai2023sigmoid,mu2022slip,bolya2025perception} are widely used in many applications~\cite{gui2024knnclip,yu2023convolutions,sain2023clip,zhou2022conditional,parelli2023clip,liu2025continual}. Among these, CLIP is a prominent representative: it consists of an image and text encoder trained to output image and text embeddings that are aligned after projection into a shared embedding space. This embedding space can be considered a normalized hypersphere~\cite{liang2022mind,wang2020understanding}. 

Several recent studies~\cite{schrodi2024two,liang2022mind,levi2025the,gandelsman2024interpreting,mistretta2025cross,shi2023towards} have analyzed critical properties of VLMs which question our understanding of the shared multi-modal embedding space and its impact on inter-modal and intra-modal tasks. \citet{liang2022mind} showed that a \textit{modality gap} exists between the image and text embeddings, even though they are trained to be aligned. This phenomenon was later attributed to information imbalance between image and text data~\cite{schrodi2024two}. 

In addition to inter-modal tasks, VLM encoders have also been repurposed as standalone encoders for tasks like image-to-image retrieval in which only the pre-trained or finetuned image encoder is used for retrieval~\cite{mistretta2025cross,kordopatis2025ilias,nara2024revisiting,nakata2022revisiting,schall2024optimizing}. In this paper, we show that CLIP intra-modal similarities are sub-optimal and propose an effective, training-free approach to adapt pre-trained VLMs for intra-modal tasks.

\minisection{Intra-modal misalignment.} Recent works~\cite{udandarao2023sus,mistretta2025cross,yi2024leveraging} investigated the \textit{intra-modal misalignment} in the CLIP embedding space. \citet{udandarao2023sus} showed that the intra-modal image-to-image and text-to-text similarities have larger means and variances than inter-modal similarities. \cite{mistretta2025cross} proposed optimization-based methods (OTI and OVI) to invert the modality of the query by crossing the modality gap and performing \textit{intra-modal tasks} using \textit{inter-modal similarities}. These methods are expensive and require many optimization steps per query. In this work, we analyze the geometric properties of the CLIP projectors and how they combine, via cosine similarity, into a single \textit{inter-modal operator}. We then exploit the spectral properties of the inter-modal operator to achieve better intra-modal alignment.

\section{Inter- and Intra-modal Operators in CLIP}
Prior work~\cite{mistretta2025cross} has empirically shown that CLIP is suboptimal on intra-modal tasks like image-to-image or text-to-text retrieval when using a single modality encoder. This limitation has been linked to the \textit{intra-modal misalignment}: CLIP's contrastive loss aligns images and text, without promoting similarity within each modality.

In this section we expand on these findings by analyzing how the CLIP training loss reinforces inter-modal alignment. We show that this alignment can be formally understood from the structure of the CLIP projection heads.

\subsection{The Role of the Projection Heads}
\label{sec:role-projection-heads}
We define the image and text encoders of CLIP as two mappings $f_{\theta}$ and $g_{\phi}$, parametrized by weights $\theta$ and $\phi$, respectively, which produce an image embedding $f_i=f_{\theta}(i)\in \mathbb{R}^{d_i}$ and a text embedding $g_t=g_{\phi}(t)\in \mathbb{R}^{d_t}$ for an image $i$ and text $t$. Our goal is to characterize the inter-modal alignment and intra-modal misalignment between the image and text encoders of CLIP. 

\minisection{The cosine similarity under projection.} CLIP maps image and text features $f_i$ and $g_t$ through two linear projector matrices $W_i \in \mathbb{R}^{d \times d_i}$ and $W_t \in \mathbb{R}^{d \times d_t}$, projecting them onto a shared  $d$-dimensional embedding space: $F(f_i)=W_i f_i \in \mathbb{R}^{d}$ and $G(g_t)=W_t g_t \in \mathbb{R}^{d}$.
The similarity between image-text pairs is computed as:
\begin{equation}
\label{eq:similarity}
    \!\!\text{sim}(f_i, g_t)=\!\! \frac{F(f_i)^{\top}\!G(g_t)}{\vert\vert F(f_i)||_2\,\! ||G(g_t)||_2}  
    \!\!=\!\!\frac{f_i^{\top} (W_i^{\top}\! W_t) g_t}{\|W_i f_i\|_2 \,\! \|W_t g_t\|_2}.
\end{equation}
This shows that CLIP always relies on the \textit{product between the two projectors} to compute cosine similarities. This product, which we define as 
\begin{equation}
  \Psi=W_i^{\top} W_t \in \mathbb{R}^{d_i \times d_t} 
\end{equation}
acts as an \emph{inter-modal operator} $\Psi: \mathbb{R}^{d_t} \rightarrow \mathbb{R}^{d_i}$ that transforms the pre-projection text feature $g_t$ into the corresponding image space. Its transpose $\Psi^{\top}$ is the reverse mapping from images to texts. Thus, $\Psi$ effectively serves as a bridge between two modalities. We will now show, via an analysis of the contrastive CLIP loss, that this operator is responsible for the \textit{inter-modal} alignment between image and text encoders during training. This analysis will also unveil a hidden \emph{intra-modal} operator that is the cause of \textit{intra-modal misalignment} in CLIP.

\minisection{Inter-modal and intra-modal operators.} The symmetric contrastive loss of CLIP is defined as:
\begin{equation}
    \mathcal{L}_{\text{CLIP}} = \frac{1}{2} (\mathcal{L}_{i \rightarrow t} + \mathcal{L}_{t \rightarrow i}  ), 
\end{equation}
where $\mathcal{L}_{i \rightarrow t}$ moves the embedding of each image $i$ toward the corresponding \textit{positive} paired text $t$, while pushing it away from all other texts in the mini-batch. The term $\mathcal{L}_{t \rightarrow i}$ performs the symmetric alignment in the opposite direction. We focus on $\mathcal{L}_{i \rightarrow t}$ for simplicity; it is defined as:
\begin{equation}
    \mathcal{L}_{i\rightarrow t} = - \log 
\frac{
    \exp\!\big( \text{sim}(f_i, g_{t}) / \tau \big)
}{
    \sum_{t'} \exp\!\big(\text{sim}(f_i, g_{t'}) / \tau \big)
}
\end{equation}
where $\tau$ is the temperature, $t$ is the  \textit{positive text} for image $i$, $t'$ indexes both the positive and the negative texts in the mini-batch, and $\text{sim}(\cdot, \cdot)$ is the cosine similarity between the image and text features as defined in \cref{eq:similarity}. To understand the effect of the inter-modal operator on the training dynamics of CLIP, we consider the gradient of the loss with respect to the pre-projection image feature $f_i$, focusing on the contribution of the pre-projection \textit{positive} text embedding $g_{t}$:
\begin{equation}
\frac{\partial \mathcal{L}_{i\rightarrow t}}{\partial f_i}
\;=\;
\frac{\partial \mathcal{L}_{i\rightarrow t}}{\partial s_{t}}
\;\frac{\partial s_{t}}{\partial f_i},
\quad
\frac{\partial \mathcal{L}_{i\rightarrow t}}{\partial s_{t}}
= \frac{1}{\tau}\big(p_{t}-1\big),
\end{equation}
with $p_{t}$ the softmax probability for the positive text $t$ and $s_t = \text{sim}(f_i, g_t)$. The gradient of the similarity for the positive image–text pair $(f_i,g_{t})$
with respect to the pre-projection image feature $f_i$ is given by:
\vspace{-0.5em}
\begin{equation}
\label{eq:grad-f}
\frac{\partial s_{t}}{\partial f_i}
= \alpha_{t, i}\,
\overbrace{W_i^{\top}W_t}^{\Psi}\,g_{t}
- s_{t}\,
\frac{\overbrace{W_i^{\top} W_i}^{\Psi_i}\,f_i}{\|W_i f_i\|_2^{2}},
\end{equation}
where $\alpha_{t, i}$ is a normalization factor. The full derivation is provided in the Supplementary Material (Sec.~\ref{sec:supp-A}).

This expression admits a clear interpretation: during training, the loss drives the image feature $f_i$ toward its paired text feature $g_{t}$, first projected into the shared space by $W_t$ and then pulled back into the image space by $W_i^{\top}$. The operator $\Psi=W_i^{\top}W_t$ therefore acts as an \textit{inter-modal operator}, responsible for aligning image and text embeddings. 
The second term involves $\Psi_i = W_i^{\top}W_i$, an \emph{intra-modal operator} which only enforces unit-norm constraints on image features but does not induce attraction between image features. Thus, during CLIP training, images interact with text via the inter-modal operator $\Psi$, encouraging inter-modal alignment. The only intra-modal interactions (via $\Psi_i$) are between an image and \textit{itself}, which do nothing to promote intra-modal alignment. Nevertheless, $\Psi_i$ implicitly defines the cosine similarity used in image-to-image retrieval or classification between image features $f_i$ and $f_{\hat{i}}$:
\begin{equation}
\label{eq:intra-modal-similarities-rev}
     \!\text{sim}(f_i, f_{\hat{i}})=\!\frac{f_{i}^{\top} (W_i^{\top} W_i) f_{\hat{i}}}{\|W_i f_i\|_2 \, \|W_i f_{\hat{i}}\|_2} \!= \!\frac{f_{i}^{\top} \Psi_i f_{\hat{i}}}{\|W_i f_i\|_2 \, \|W_i f_{\hat{i}}\|_2},\!
\end{equation}
even though it is not trained to align images, making it suboptimal for modeling image-to-image relationships.

\begin{figure}
    \centering
\includegraphics[width=0.95\linewidth]{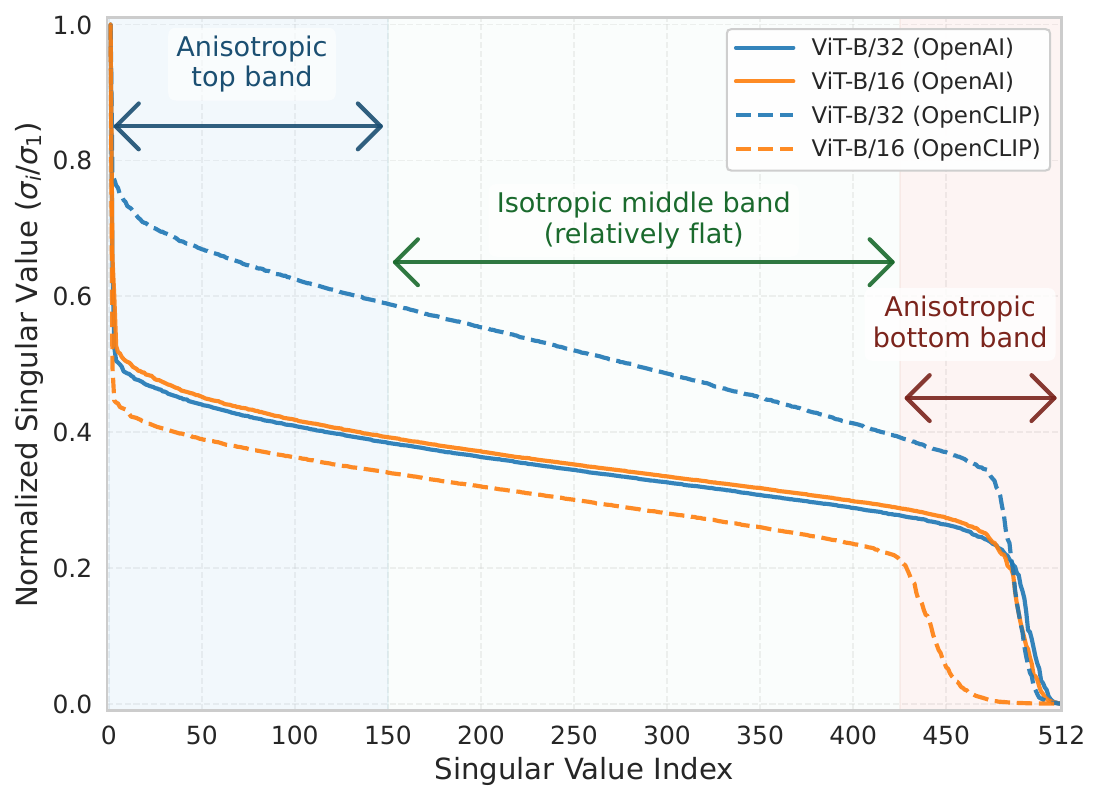}
\caption{Spectra of the inter-modal operator $\Psi=W_i^{\top} W_t$ for CLIP ViT-B/16 and ViT-B/32 with OpenAI and DataComp (OpenCLIP) pre-training. Despite variations across models, all spectra show pronounced anisotropy in the extreme top and bottom singular directions, while staying relatively flat in the middle band.}
\label{fig:singular-values}
\end{figure}

\subsection{Spectral Analysis of the Inter-modal Operator}
From the previous discussion, $\Psi = W_i^{\top} W_t$ and its transpose $\Psi^{\top} = W_t^{\top} W_i$ define a pair of linear mappings that connect the pre-projection spaces of the two modalities. To analyze the distortion introduced by these operators, we compute the Singular Value Decomposition (SVD) of $\Psi$:
\begin{equation}
 \Psi = U \Sigma V^{\top}.
\end{equation}
where $U\in \mathbb{R}^{d_i\times d}$ contains the left singular vectors spanning the output (image) space, and $V\in \mathbb{R}^{d_t \times d}$ contains the right singular vectors spanning the input (text) space. The diagonal matrix $\Sigma \in \mathbb{R}^{d \times d}$ contains the singular values, which quantify how directions in one modality are stretched or compressed when mapped to the other through $\Psi$ or $\Psi^{\top}$. It suffices to analyze $\Psi$, as $\Psi^{\top}$ shares the same singular values; the SVD of $\Psi^{\top}$ simply exchanges the roles of the input and output spaces without altering the spectrum.

In \cref{fig:singular-values} we see that the singular values are relatively flat across a broad central region, while exhibiting strong anisotropy in both the top and bottom parts of the spectrum. The extensive, relatively flat region in the middle reveals an approximately \textit{isotropic} subspace in which features can be transferred across modalities with minimal distortion. This suggests the presence of a shared semantic subspace \textit{directly identifiable from the projector weights themselves}. Conversely, the  top and bottom directions capture modality-specific variations which we analyze in the next section.  

Building on these observations, we now investigate whether this shared semantic subspace learned by CLIP can be exploited to improve intra-modal alignment when CLIP is used in intra-modal settings.

\section{IsoCLIP: Aligning Projector Weights by Decomposing the Inter-modal Operator} 
\label{sec:main-4}
\begin{figure*}[t]
    \centering
    \includegraphics[width=0.98\linewidth]
    {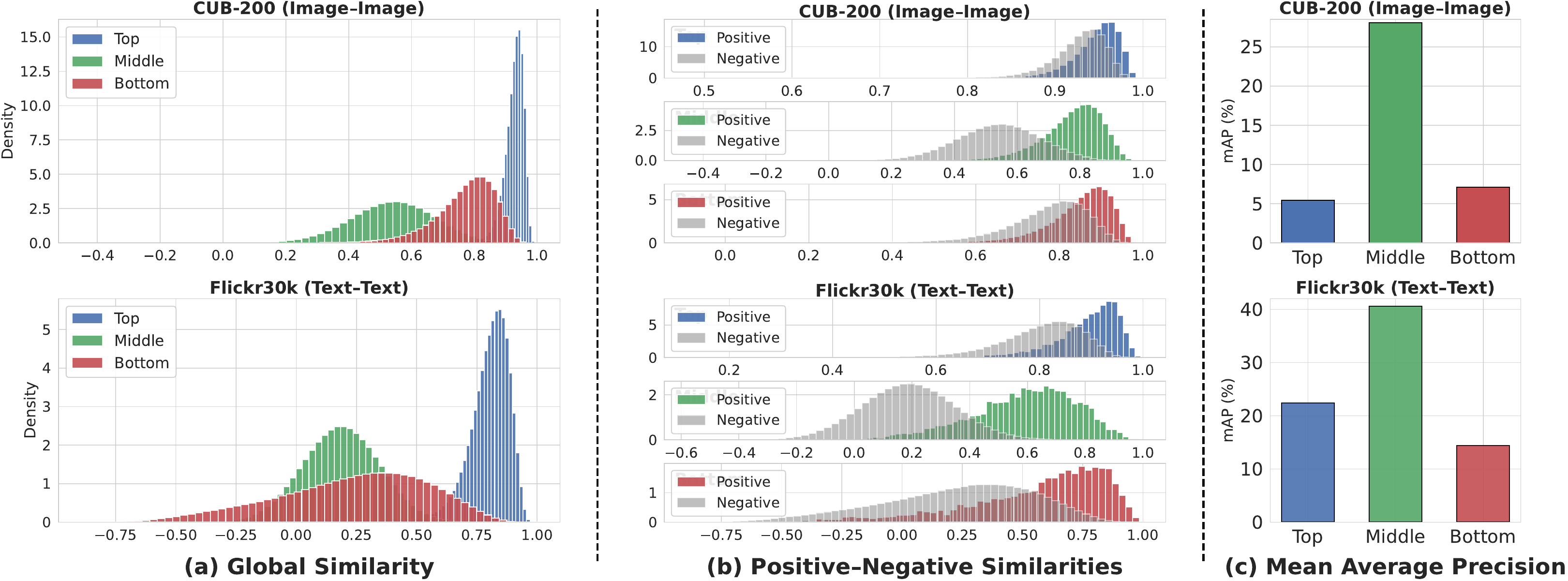}
    \caption{
    Investigation of different regions of the spectrum (top 50, middle 50, and bottom 50 directions) of the inter-modal operator $\Psi$ for aligning the CLIP Projector weights, as defined in \cref{proj:equation}, utilizing the ViT-B/16 model for both image-to-image and text-to-text retrieval tasks. \textbf{(a)} Analysis of cosine similarity distributions showing very high similarities for the top band, and more well-behaved distributions for middle and bottom bands. \textbf{(b)} Overlap between cosine similarities of positive and negative pairs showing better separation for the middle band but highly overlapping distributions for top and bottom bands, implying higher intra-modal misalignment. \textbf{(c)} Performance comparison showing far superior performance using the well-aligned middle band compared to top and bottom bands.}
    \label{fig:method_analysis}
\end{figure*}
 
The analysis in Sec.~\ref{sec:role-projection-heads} suggests that intra-modal similarities in CLIP rely on a geometry induced by the intra-modal operator $\Psi_i=W_i^{\top}W_i$ which is not optimized during training (Eqs.~\eqref{eq:grad-f}, \eqref{eq:intra-modal-similarities-rev}). We therefore investigate whether the inter-modal operator $\Psi$, optimized during training, can reveal shared semantic directions across modalities for improved intra-modal similarity.
We decompose the inter-modal operator as $\Psi= U \Sigma V^{\top}$, where $U, V$ span the image and text spaces and $\Sigma$ contains singular values in decreasing order. We then restrict $\Psi$ to the relatively flat central region of its spectrum $\Sigma$ by retaining only the corresponding singular directions. Formally, we select the paired subspaces:   
\begin{equation}
\label{eq:top-bottom}
\begin{aligned}
    \mathcal{S}_U &=\text{span}\{u_j  | j \in [k_t, r-k_b]\} \mbox{ and}     \\
    \mathcal{S}_V &=\text{span}\{v_j | j \in [k_t, r-k_b]\},
\end{aligned}
\end{equation}
where $u_j$ and $v_j$ identify the $j$-th column of $U$ and $V$ respectively, $r$ is the rank of $\Psi$, and $k_t$ and $k_b$ delimit the range of approximately isotropic singular values.
Projecting the image and text projectors onto these subspaces yields:
\begin{equation}
\label{proj:equation}
    \widehat{W}_i = W_i U_{\mathcal{S}_U} U_{\mathcal{S}_U}^{\top}, \quad     \widehat{W}_t = W_t V_{\mathcal{S}_V} V_{\mathcal{S}_V}^{\top}.
\end{equation}
This projection defines our approach, IsoCLIP.
This operation, via the orthogonal projectors $U_{\mathcal{S}_U} U_{\mathcal{S}_U}^{\top}$  and $V_{\mathcal{S}_V} V_{\mathcal{S}_V}^{\top}$, restricts $W_i$ and $W_t$ to the middle-band of the inter-modal operator $\Psi$, filtering out the anisotropic top and bottom directions and aligning both mappings to a shared semantic space. The resulting projectors $\widehat{W}_i$ and $\widehat{W}_t$ therefore operate in a common basis, corresponding to the subspace where image and text representations exhibit minimal distortion. For intra-modal tasks, the IsoCLIP projectors are used to compute the cosine similarities within the isotropic subspace. For image-to-image retrieval or classification, the similarities between images $i$ and $\hat{i}$ are computed as:
\begin{equation}
    \text{sim}(f_i, f_{\hat{i}})= \frac{f_{i}^{\top} (\widehat{W}_i^{\top} \widehat{W}_i) f_{\hat{i}}}{\|\widehat{W}_i f_i\|_2 \, \|\widehat{W}_i f_{\hat{i}}\|_2},
\end{equation}
and analogously for text-to-text retrieval using $\widehat{W}_t$. Using $\widehat{W}_i$ for intra-modal similarity flattens the spectrum of $W_i^{\top}W_i$ (Eq.~\ref{eq:intra-modal-similarities-rev}), spreading similarity across more directions, improving positive–negative separation, and raising mAP. See Sec.~\ref{sec:supp-B} of the Supplementary Material for details.

\minisection{Semantics of the middle band of the spectrum.} The singular directions associated with the middle band of the spectrum $[k_t, r-k_b]$ (see \cref{eq:top-bottom}) contain the shared semantic directions that support CLIP's inter-modal alignment. To investigate this hypothesis, we isolate three spectral regions in ViT-B/16 -- the \textit{top interval} $[0,k_t]$, the \textit{bottom interval} $[r-k_b, r]$, and a \textit{middle band} defined symmetrically around midpoint of the spectrum, $[r/2 - k_t/2, r/2 + k_b/2]$, where $r$ is the rank and $k_t=k_b=50$. This choice is motivated by the singular value profile in \cref{fig:singular-values}, which shows sharp slope changes in the first and last $\sim$50 components, while the central region is comparatively flat and thus expected to encode more semantic structure. Here we use 50 dimensions for the bands for analysis. See~\cref{sec:ablation} for a discussion of how optimal values are selected in all experiments.

We restrict the projection matrix of the image and text encoder to each subspace (\cref{proj:equation}) and analyze how the cosine similarities between query and gallery features change when only the top, middle, or bottom directions are used. ~\cref{fig:method_analysis} summarizes the effect on global and positive-negative similarity distributions, and retrieval performance for both image-to-image and text-to-text retrieval. We see that:
\begin{itemize}
\item \textbf{Top directions} inflate cosine similarities (\cref{fig:method_analysis}a) but mix positive and negative pairs (\cref{fig:method_analysis}b), which leads to poor retrieval performance (\cref{fig:method_analysis}c).
\item \textbf{Bottom directions} produce less inflated similarities (\cref{fig:method_analysis}a) but still fail to separate positives from negatives (\cref{fig:method_analysis}b), resulting in poor performance (\cref{fig:method_analysis}c).
\item \textbf{Middle directions}, in contrast, yield well-behaved similarity distributions (\cref{fig:method_analysis}a, \ref{fig:method_analysis}b) and consistently superior retrieval accuracy across modalities (\cref{fig:method_analysis}c).
\end{itemize}
An interesting observation is the \textit{asymmetric effect across modalities}: top directions are more detrimental to image retrieval than to text retrieval, while bottom directions show the opposite trend (\cref{fig:method_analysis}c). This pattern suggests that the spectral extremes capture \textit{modality-specific variation}, with the top directions predominantly text-specific and the bottom ones predominantly image-specific. Both extremes are detrimental for intra-modal tasks, whereas the middle band encodes the semantic subspace that both modalities rely on and is more discriminative for intra-modal tasks. See the Supplementary Material (Sec.~\ref{sec:supp-C}) for additional evidence.

\minisection{Extension to non-linear projector heads.} From Eq.~\ref{proj:equation} is clear that IsoCLIP applies directly to CLIP-style models with linear projection heads. For models such as SigLIP2~\cite{tschannen2025siglip2multilingualvisionlanguage}, which employ a non-linear image-projection, IsoCLIP can be extended via a first-order linearization of the projection head~\cite{Haoqi_2024_ECCV}. In Sec.~\ref{sec:supp-linearization} of the Supplementary Materials we provide details of the proposed linearization.

\begin{table*}[h]
\centering
\setlength{\tabcolsep}{3pt}
\renewcommand{\arraystretch}{1.05}
\caption{\textbf{Image-to-image retrieval} performance for multiple CLIP models on 13 datasets. We also report the query latency for all methods.}
\label{tab:iso_comparison_imagetoimage}
\resizebox{0.96\textwidth}{!}{
\begin{tabular}{lccccccccccccccccc}
\toprule
\textbf{Method} & \textbf{Intra-modal} &\textbf{Backbone}  & \textbf{Latency (ms)} & \textbf{Caltech} & \textbf{CUB} & \textbf{ROxford} & \textbf{RParis} & \textbf{Cars} & \textbf{Pets} & \textbf{Flowers} & \textbf{Aircraft} & \textbf{DTD} & \textbf{EuroSAT} & \textbf{Food101} & \textbf{SUN397}  & \textbf{UCF101} & \textbf{Avg} \\
\midrule

Image-Image & \Checkmark & \multirow{3}{*}{ViT-B/32} & 7 $\pm$ 1
& 77.1 & 22.9 & 42.6 & 67.9 & 24.6 & 30.5 & 62.0 & \underline{14.5} & 28.1 & \underline{47.9} & 32.3 & 34.3 & 47.1 & 40.9 \\
OTI (I$\rightarrow$T) &  \XSolidBrush &  &  1879 $\pm$ 35  
& \underline{79.9} & \underline{24.6} & \underline{43.0} & \underline{70.3} & \underline{28.0} & \underline{37.5} & \underline{62.6} & 14.4 & \textbf{31.9} & 47.2 & \underline{34.7} & \underline{36.3} &  \textbf{48.6} & \underline{43.0} \\
\hlrow IsoCLIP & \Checkmark & & 7 $\pm$ 1
& \textbf{80.8} & \textbf{27.0} & \textbf{47.2} & \textbf{73.8} & \textbf{30.0} & \textbf{40.8} & \textbf{66.5} & \textbf{14.9} &\underline{30.9} & \textbf{51.5} & \textbf{38.0} & \textbf{36.4} & \underline{48.4} & \textbf{45.1} \\
\midrule

Image-Image & \Checkmark & \multirow{3}{*}{ViT-B/16} & 6 $\pm$ 1
& 80.6 & 31.6 & 46.6 & 75.3 & 31.0 & 36.3 & 70.8 & 19.0 & 30.7 & \underline{51.2} & 42.8 & 35.9 & 49.8 & 46.3 \\
OTI (I$\rightarrow$T) & \XSolidBrush  & &  1856 $\pm$ 56
& \underline{83.5} & \underline{33.9} & \underline{49.9} & \underline{77.4} & \underline{37.2} & \underline{42.9} & \underline{72.8} & \underline{20.1} & \underline{35.1} & 50.5 & \underline{47.5} & \textbf{38.7} & \underline{52.6} & \underline{49.4} \\
\hlrow IsoCLIP & \Checkmark & &  6 $\pm$ 1
& \textbf{85.0} & \textbf{38.6} & \textbf{51.8} & \textbf{82.0} & \textbf{41.2} & \textbf{50.7} & \textbf{77.4} & \textbf{20.5} & \textbf{36.0} & \textbf{55.6} & \textbf{53.5} & \underline{38.5} & \textbf{55.4} & \textbf{52.8} \\
\midrule

Image-Image & \Checkmark & \multirow{3}{*}{ViT-L/14} & 11 $\pm$ 1
& 83.2 & 43.0 & 57.5 & 76.9 & 43.3 & 47.3 & 84.0 & 25.8 & 34.1 & \underline{59.0} & 53.0 & 39.1 & 60.0 & 54.3 \\
OTI (I$\rightarrow$T) & \XSolidBrush & & 1872 $\pm$ 91
&  \textbf{87.3} & \underline{47.1} & \underline{62.4} & \underline{77.1} & \underline{50.5} & \underline{56.0} & \underline{86.0} & \underline{27.1} & \underline{37.7} & 56.3 & \underline{55.9} & \textbf{43.5} & \textbf{62.8} & \underline{57.7} \\
\hlrow IsoCLIP & \Checkmark & & 11 $\pm$ 1
&  \underline{87.0} & \textbf{52.2} & \textbf{66.4} & \textbf{81.4} & \textbf{56.4} & \textbf{63.5} & \textbf{88.2} & \textbf{28.2} & \textbf{39.0} & \textbf{61.6} & \textbf{62.9} & \underline{41.0}& \underline{61.9} & \textbf{60.7} \\
\midrule

Image-Image & \Checkmark & \multirow{3}{*}{ViT-B/16-open} & 6 $\pm$ 1
& 85.7 & 42.8 & 65.3 & 83.2 & 55.8 & 50.4 & 84.6 & 23.1 & 39.9 & 57.8 & 51.1 & \underline{39.5} &  52.9 & 56.3 \\
OTI (I$\rightarrow$T) & \XSolidBrush & & 1836 $\pm$ 83
& \underline{85.8} & \underline{45.1} & \textbf{69.5} & \textbf{85.8} & \underline{60.5} & \underline{56.5} & \underline{85.2} & \underline{23.4} & \textbf{43.1} & \textbf{58.8} & \underline{54.4} & \textbf{40.8} &  \textbf{54.1} & \textbf{58.7} \\
\hlrow IsoCLIP &  \Checkmark & & 6 $\pm$ 1
& \textbf{87.6} & \textbf{45.9}& \underline{67.3} & \underline{85.0} & \textbf{60.7} & \textbf{57.8} & \textbf{85.8} & \textbf{23.5} & \underline{42.5} & \underline{58.6} & \textbf{54.7} & 39.3 & \underline{53.4} & \underline{58.6} \\
\midrule

Image-Image & \Checkmark & \multirow{3}{*}{PE-Core-B-16} & 8 $\pm$ 1
& 89.1 & 48.9 & 61.6 & 83.4 & 53.4 & 57.1 & 85.0 & 31.7 & 39.0 & 55.1 & 53.4 & 43.9 & 60.3 & 58.6 \\
OTI (I$\rightarrow$T) & \XSolidBrush & & 3252 $\pm$ 35
& \underline{90.5} & \underline{51.5} & \underline{65.7} & \underline{85.2} & \underline{60.2} & \underline{62.9} & \underline{87.2} & \underline{32.7} & \textbf{43.5} & \textbf{56.9} & \underline{55.8} & \underline{44.9} & \textbf{63.0} & \underline{61.5} \\
\hlrow IsoCLIP & \Checkmark & & 8 $\pm$ 1
& \textbf{91.6} & \textbf{53.0} &\textbf{67.0} & \textbf{86.5} & \textbf{62.3} & \textbf{68.2} & \textbf{88.0} & \textbf{34.9} & \underline{43.1} & \underline{56.3} & \textbf{57.6} & \textbf{45.2} & \underline{62.1} & \textbf{62.8} \\
\bottomrule
\end{tabular}
}
\end{table*}

\begin{table}
\setlength{\tabcolsep}{2pt} 
\caption{\textbf{Text-to-text retrieval} performance for multiple CLIP models on three datasets.}
\label{tab:iso_text_comparison} 
\resizebox{\columnwidth}{!}{
\begin{tabular}{lccccccc}
\toprule
\textbf{Method} & \textbf{Intra-modal} & \textbf{Backbone} & \textbf{Latency (ms)}
& \textbf{COCO} & \textbf{Flickr30k} & \textbf{nocaps} & \textbf{Avg} \\
\midrule

Text-Text & \Checkmark & \multirow{3}{*}{ViT-B/32} & 6 $\pm$ 1
& 26.2 & 51.7 & 35.1 & 37.7 \\
OVI (T$\rightarrow$I) & \XSolidBrush & &  11549 $\pm$ 192
& \underline{28.3} & \underline{54.8} & \underline{38.8} & \underline{40.6} \\
\hlrow IsoCLIP & \Checkmark &  & 6 $\pm$ 1
& \textbf{29.1} & \textbf{56.8} & \textbf{39.6} & \textbf{41.9} \\
\midrule

Text-Text & \Checkmark &  \multirow{3}{*}{ViT-B/16} & 6 $\pm$ 1
& 26.1 & 50.7 & 35.1 & 37.3 \\
OVI (T$\rightarrow$I) & \XSolidBrush & &  11929 $\pm$ 125 
& \underline{28.2} &\underline{53.1} & \underline{38.3} & \underline{39.9} \\
\hlrow IsoCLIP & \Checkmark & &  6 $\pm$ 1
& \textbf{28.9} & \textbf{55.2} & \textbf{39.4} & \textbf{41.2} \\
\midrule

Text-Text & \Checkmark & \multirow{3}{*}{ViT-L/14} & 6 $\pm$ 1
& 26.7 & 52.3 & 35.7 & 38.2 \\
OVI (T$\rightarrow$I) & \XSolidBrush &  & 21366 $\pm$ 33
& \underline{29.4} & \underline{54.9} & \underline{39.5} & \underline{41.3} \\
\hlrow IsoCLIP & \Checkmark & &  6 $\pm$ 1
& \textbf{30.1} & \textbf{58.6} & \textbf{40.4} & \textbf{43.0} \\
\midrule

Text-Text & \Checkmark &  \multirow{3}{*}{ViT-B/16-open} & 6 $\pm$ 1
& 29.8 & 57.0 & 40.0 & 42.3 \\
OVI (T$\rightarrow$I) & \XSolidBrush & & 11822 $\pm$ 169
& \textbf{31.9} & \underline{60.2} & \textbf{43.7} & \textbf{45.3} \\
\hlrow IsoCLIP & \Checkmark &  & 6 $\pm$ 1
& \textbf{31.9} & \textbf{60.8} & \underline{43.3} & \textbf{45.3} \\
\bottomrule
\end{tabular}
}
\end{table}

\section{Experimental Results}
\label{sec:experiments}

In this section we report on experiments evaluating IsoCLIP in a range of intra-modal settings, employing several VLMs, and on a diverse selection of benchmark datasets.

\subsection{Experimental Settings}

We evaluate our method on a wide range of datasets and models covering both \textit{image-to-image} and \textit{text-to-text} retrieval, following the protocol of \citet{mistretta2025cross}. 

\minisection{Datasets and metrics.} We evaluate image-to-image retrieval and image classification on a wide range of datasets, including fine-grained, scene-level, and object-centric benchmarks. For text-to-text retrieval, we use three caption datasets following \citet{mistretta2025cross}. 
See Sec.~\ref{sec:supp-E} of the Supplementary Material for full dataset details. For retrieval experiments, we report the \textit{mean average precision} (mAP), while for the image classification performance, we report the \textit{accuracy}. We also report the time taken to encode a single query using the three methods (averaged over 20 runs) as the latency in milliseconds.
 
\minisection{Backbones.} We conduct experiments using multiple CLIP variants with different backbones and pre-training datasets. Specifically, we consider \textit{OpenAI CLIP} models with ViT-B/32, ViT-B/16, and ViT-L/14 image backbones, as well as \textit{OpenCLIP} models (ViT-B/16-open) trained on the DataComp dataset. To further explore vision-language models distinct from CLIP, we use the recent \textit{Perception-Encoder} (PE) model~\cite{bolya2025perception} pre-trained on the PE Video Dataset~\cite{bolya2025perception}, allowing us to analyze generalization across diverse vision–language representations. In the Supplementary Material (Sec.~\ref{sec:supp-F}) we extend the evaluation to additional models, including ViT-B/32-open, EVA-02~\cite{FANG2024105171}, and SigLIP2 B/16.

\minisection{Middle band selection.}  Since architectures vary in both the number of singular directions and shape of their spectra (\cref{fig:singular-values}), the optimal $k_t$ and $k_b$ differ across models. We empirically select the top \(k_t\) and bottom \(k_b\) spectral components to remove, as defined in \cref{eq:top-bottom}, based on a single dataset for all backbones. For image-based tasks we validate these hyperparameters on image-to-image retrieval using Caltech101, a generic object recognition dataset. These parameters are used for all other datasets in image retrieval and classification experiments. For text-to-text retrieval, we validate these hyperparameters on COCO and use them for all other text datasets. In~\cref{sec:ablation} we discuss the impact of hyperparameter selection.

\begin{table*}[h]
\centering
\setlength{\tabcolsep}{3pt}
\renewcommand{\arraystretch}{1.05}
\caption{\textbf{Image classification} performance for multiple CLIP models on 10 datasets. We compare IsoCLIP  with intra-modal NCM classification. Zero-shot results are reported only for reference, as they do not require training images to compute class prototypes.}
\label{tab:iso_comparison_classification}
\resizebox{0.96\textwidth}{!}{
\begin{tabular}{lcccccccccccccc}
\toprule
\textbf{Method} & \textbf{Intra-modal} & \textbf{Classifier} & \textbf{Backbone} & \textbf{Caltech} & \textbf{Cars} & \textbf{Pets} & \textbf{Flowers} & \textbf{Aircraft} & \textbf{DTD} & \textbf{EuroSAT} & \textbf{Food101} & \textbf{SUN397}  & \textbf{UCF101} & \textbf{Average} \\
\toprule
Image-Text & \XSolidBrush & Zero-Shot & \multirow{3}{*}{ViT-B/32} & 91.2 & 60.4 & \textbf{87.5} & 67.0 & 19.1 & 43.6 & 45.2 & \textbf{80.5} & 62.0  & 62.0 & 61.9 \\
Image-Image &  \Checkmark & NCM  &  & \underline{91.9} & \underline{67.0} & 77.1 & \underline{93.7} & \underline{34.5} & \underline{64.4} & \underline{77.7} & 77.3 & \underline{70.2} & \underline{76.6} & \underline{73.0} \\
\hlrow IsoCLIP  & \Checkmark & NCM  &  & \textbf{93.0} & \textbf{73.0} & \underline{84.0} & \textbf{93.8} & \textbf{35.0} & \textbf{66.8} & \textbf{80.6} & \underline{78.6} & \textbf{70.5} & \textbf{77.8} & \textbf{75.3} \\
\midrule
Image-Text & \XSolidBrush & Zero-Shot & \multirow{3}{*}{ViT-B/16} & 92.9 & 65.3 & \underline{89.1} & 71.4 & 24.8 & 44.4 & 47.8 & \textbf{86.1} & 62.6 & 66.7 & 65.1 \\
Image-Image & \Checkmark & NCM  &  & \underline{94.1} & \underline{74.8} & 84.2 & \underline{96.2} & \underline{43.6} & \underline{65.8} & \underline{79.2} & 84.0 & \textbf{72.1} & \underline{80.2} & \underline{77.4} \\
\hlrow IsoCLIP & \Checkmark & NCM  & &  \textbf{95.4} & \textbf{81.2} & \textbf{90.9} & \textbf{97.1} & \textbf{45.6} & \textbf{68.4} & \textbf{82.6} & \underline{85.6} & \underline{72.0} & \textbf{81.7} & \textbf{80.1} \\
\midrule
Image-Text & \XSolidBrush & Zero-Shot & \multirow{3}{*}{ViT-L/14}  & 94.8 & 76.8 & \underline{93.6} & 79.3 & 32.5 & 53.1 & 58.1 & \textbf{91.0} & 67.6 & 74.2 & 72.1 \\
Image-Image & \Checkmark & NCM  &  & \underline{96.8} & \underline{83.7} & 91.8 & \underline{98.8} & \underline{55.5} & \underline{70.4} & \underline{84.7} & 90.0 & \textbf{76.7} & \underline{85.9} & \underline{83.4} \\
\hlrow IsoCLIP  & \Checkmark & NCM  &  & \textbf{97.2} & \textbf{87.8} & \textbf{94.4} & \textbf{99.0} & \textbf{58.0} & \textbf{72.9} & \textbf{88.5} & \underline{90.8} & \underline{75.8} & \textbf{86.9} & \textbf{85.1} \\
\bottomrule
\end{tabular}
}
\end{table*}

\subsection{Comparison with Standard and Inversion-based Approaches}
In this section we evaluate IsoCLIP on image-to-image retrieval and text-to-text retrieval.

\minisection{Image-to-image retrieval.} We compare IsoCLIP against standard image-to-image retrieval (Image-Image) using the vision encoder only, and against the existing textual inversion-based approach (OTI). OTI maps the query image features into the text space by performing 150 steps of optimization and then retrieves from the image gallery, following \cite{mistretta2025cross}. By inverting an image into a text embedding, OTI compares the inverted pseudo-text query against the gallery image embeddings, resulting in an inter-modal comparison and thereby circumventing intra-modal misalignment.

In \cref{tab:iso_comparison_imagetoimage} we see that IsoCLIP significantly improves the performance over standard image-to-image retrieval by significant margins across all backbones (6.5\% on ViT-B/16, and 4.2\% on Perception-Encoder on average). IsoCLIP outperforms OTI with significantly less query latency on all backbones except ViT-B/16-open where it performs similar to OTI. The improved performance of IsoCLIP compared to standard CLIP Image-Image confirms the potential of using the modality-aligned directions for intra-modal tasks.

\minisection{Text-to-text retrieval.} We compare IsoCLIP with standard text-to-text retrieval using the text encoder only (Text-Text), and against the existing visual inversion-based approach (OVI). OVI maps the query text feature into the image space by performing 1000 optimization steps and then retrieves from the text gallery~\cite{mistretta2025cross}. Converting a text query into an image embedding, OVI results in an inter-modal comparison and circumvents intra-modal misalignment.

We report in~\cref{tab:iso_text_comparison} the comparison of IsoCLIP with standard text-to-text retrieval and OVI. We observe that IsoCLIP outperforms standard retrieval significantly on all datasets (3.9\% using ViT-B/16, 4.8\% using ViT-L/14 on average). Despite having negligible latency compared to OVI, IsoCLIP performs similar or better than OVI in most settings.

\begin{table*}[h]
\centering
\setlength{\tabcolsep}{3pt}
\renewcommand{\arraystretch}{1.05}
\caption{\textbf{Ablation Study.} We compare IsoCLIP against using pre-projection image features for retrieval (Image–Image [Pre]) and against whitening the CLIP image projection weights ($W_i^{\text{white}}$). } 
\label{tab:ablation_studies}
\resizebox{0.96\textwidth}{!}{
\begin{tabular}{llcccccccccccccc}
\toprule
\textbf{Method} & \textbf{Backbone} & \textbf{Caltech} & \textbf{CUB} &  \textbf{ROxford} & \textbf{RParis} & \textbf{Cars} & \textbf{Pets} & \textbf{Flowers} & \textbf{Aircraft} & \textbf{DTD} & \textbf{EuroSAT} & \textbf{Food101} & \textbf{SUN397} & \textbf{UCF101} & \textbf{Average} \\
\toprule
Image-Image & \multirow{4}{*}{ViT-B/16}  & 80.6 & 31.6 & 46.6 & 75.3 & 31.0 & 36.3 & 70.8 & 19.0 & 30.7 & 51.2 & 42.8 & 35.9 & 49.8 & 46.3 \\
Image-Image [Pre] &  & 81.6 & 32.7 & \underline{49.2} & 77.8 & 32.0 & 39.0 & 72.4 & 19.4 & 32.5 & \underline{54.5} & 45.3 & \textbf{38.8} & \underline{54.2} & 48.4 \\
  $W_i^{\text{white}}$ &  & \underline{82.1} & \underline{34.4} & 48.5 & \underline{78.6} & \underline{34.2} & \underline{40.8} & \underline{73.7} & \underline{19.5} & \underline{32.9} & 54.3 & \underline{46.5} & 38.2 & 53.9 & \underline{49.0} \\
 
\hlrow IsoCLIP &  & \textbf{85.0} & \textbf{38.6} & \textbf{51.8} & \textbf{82.0} & \textbf{41.2} & \textbf{50.7} & \textbf{77.4} & \textbf{20.5} & \textbf{36.0} & \textbf{55.6} & \textbf{53.5} & \underline{38.5} & \textbf{55.4} & \textbf{52.8} \\
 \midrule
 
 Image-Image & \multirow{4}{*}{ViT-L/14} 
& 83.2 & 43.0 & 57.5 & 76.9 & 43.3 & 47.3 & 84.0 & 25.8 & 34.1 & 59.0 & 53.0 & 39.1 & 60.0 & 54.3 \\
Image-Image [Pre] &  & 85.4 & 42.6 & 59.5 & \underline{78.1} & 41.9 & 48.0 & 84.7 & 25.9 & 35.3 & \textbf{61.7} & 54.6 & \textbf{42.0} & \textbf{61.9} & 55.5 \\
 $W_i^{\text{white}}$&  & \underline{85.6} & \underline{44.0} & \underline{60.1} & 77.6 & \underline{44.2} & \underline{49.2} & \underline{85.3} & \underline{26.1} & \underline{35.7} & 61.3 & \underline{55.1} & \underline{41.6} & \textbf{61.9} & \underline{56.0} \\
 
\hlrow IsoCLIP &  & \textbf{87.0} & \textbf{52.2} & \textbf{66.4} & \textbf{81.4} & \textbf{56.4} & \textbf{63.5} & \textbf{88.2} & \textbf{28.2} & \textbf{39.0} & \underline{61.6} & \textbf{62.9} &  41.0 & \textbf{61.9} & \textbf{60.7} \\
 \bottomrule
\end{tabular}}
\end{table*}

\subsection{Analysis and Ablations}
\label{sec:ablation}

\minisection{Analysis on Image classification.} Existing works~\cite{udandarao2023sus,zhang2022tip,huang2024lp++,zhang2024dual} use intra-modal comparisons alongside textual information for CLIP-based image classification. To evaluate IsoCLIP in an intra-modal classification setting, we adopt the Nearest Class Mean (NCM) classifier~\cite{guerriero2018deepncm:}, which classifies images by their proximity to class prototypes computed from all training data. 

Specifically, we compute the mean of the pre-projection image features for each class and then project them to the shared embedding space using the IsoCLIP image projector. These projected class means are used as prototypes for classification instead of the text embeddings used for zero-shot classification. At inference time, the test image embeddings generated using the IsoCLIP image projector will be assigned to the class whose prototype has the highest cosine similarity to the test image.

We report in~\cref{tab:iso_comparison_classification} the performance of IsoCLIP and compare to standard image-to-image classification using NCM classifiers. The NCM classifier significantly outperforms the inter-modal zero-shot classifier (having access only to text class names) on most datasets, as it leverages the entire training set of images to compute the prototypes. On an average, IsoCLIP outperforms standard image-to-image classification significantly across three different backbones. This implies that IsoCLIP could also be exploited for classification tasks involving intra-modal comparisons.
\begin{figure}
\centering
\includegraphics[width=\linewidth]{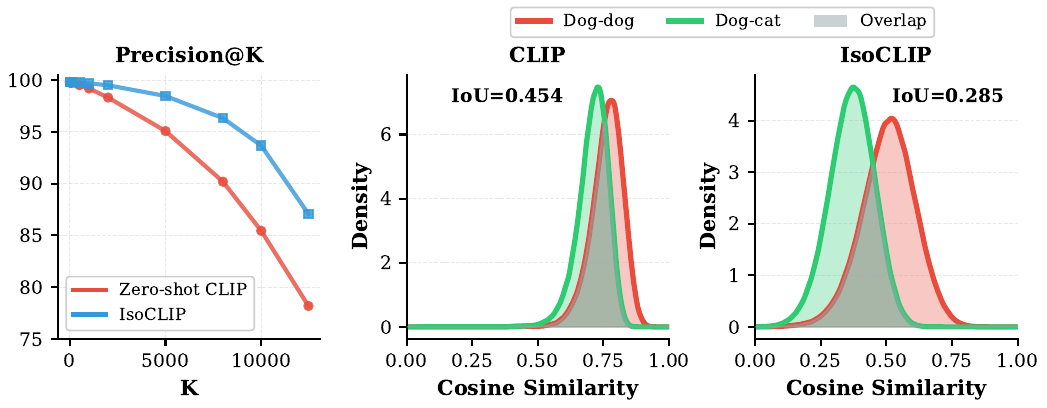}
\caption{Analysis on \textit{Dogs vs. Cats}~\cite{elson2007asirra}. (left) IsoCLIP achieves higher precision than CLIP as the number of retrieved dog images $K$ increases. (center) CLIP shows significant overlap between intra-class (dog-dog) and inter-class (dog-cat) similarities due to intra-modal misalignment. (right) IsoCLIP reduces this overlap, making image-image similarities more discriminative.}
\vspace{-10pt}
\label{fig:intra-modal-analysis}
\end{figure}

\minisection{Analysis of intra-modal misalignment.} To study how much IsoCLIP improves intra-modal alignment in a controlled and isolated setting, we evaluate on the \textit{Dogs vs. Cats} dataset which contains 25K images~\cite{elson2007asirra}. Following the preprocessing from~\cite{mistretta2025cross}, we remove dog images that are closer to ``A photo of a cat'' than to ``A photo of a dog''. This gives us a perfect inter-modally aligned subset of images, which ensures that an improvement in image-to-image retrieval performance arises only from mitigation of intra-modal misalignment.

We perform image-to-image retrieval with ViT-B/16 using dog images as queries and cats and dogs as gallery. In~\cref{fig:intra-modal-analysis} we show that IsoCLIP outperforms CLIP when varying the number of retrieved samples $K$ in terms of Precision@K. We demonstrate that the similarities between positives (Dog-to-Dog) and between negatives (Dog-to-Cat) overlap more in the CLIP embedding space (IoU=0.464) compared to IsoCLIP (IoU=0.293) -- a clear visualization of intra-modal alignment induced by IsoCLIP.

\minisection{Ablation studies.} We evaluate two additional strategies to better understand the contribution of the image projection head to intra-modal misalignment (see~\cref{sec:role-projection-heads}). 
Since intra-modal misalignment clearly manifests when using post-projection features $W_if_i$ (Image-Image), we also evaluate retrieval performance using the \textit{raw pre-projection features} $f_i$ (Image-Image [Pre]). This ablation tests whether the projection head itself is responsible for the suboptimal performance in intra-modal similarity comparisons. We also evaluate a whitened image projector obtained by decomposing $W_i = U_i\Sigma_iV_i^{\top}$ and constructing $W_i^{\text{white}}=U_iV_i^{\top}$. This flattens the spectrum of $W_i$, while preserving its directions, allowing us to test whether a uniform spectrum is sufficient to mitigate intra-modal misalignment.

\cref{tab:ablation_studies} shows that both ablations yield significant improvement over the standard Image-Image baseline: $+2.1$ mAP (Image-Image [Pre]) and $+2.7$ mAP ($W_i^{\text{white}}$) on ViT-B16, and $+1.2$ and $+2.7$ respectively on ViT-L/14. These improvements indicate that the projector introduces significant anisotropy that harms intra-modal retrieval.

IsoCLIP achieves larger gains than both alternatives. Since the isotropic middle band of $\Psi$ captures the semantic directions underlying image-text alignment, restricting the projector to this subspace is more effective than either flattening the spectrum (whitening) or bypassing the projection head (Image-Image [Pre]). See Supplementary Material (Sec.~\ref{sec:supp-F}) for results on other models. 

\begin{figure}
    \centering
    \resizebox{\columnwidth}{!}{
    {\includegraphics[width=0.49\textwidth]{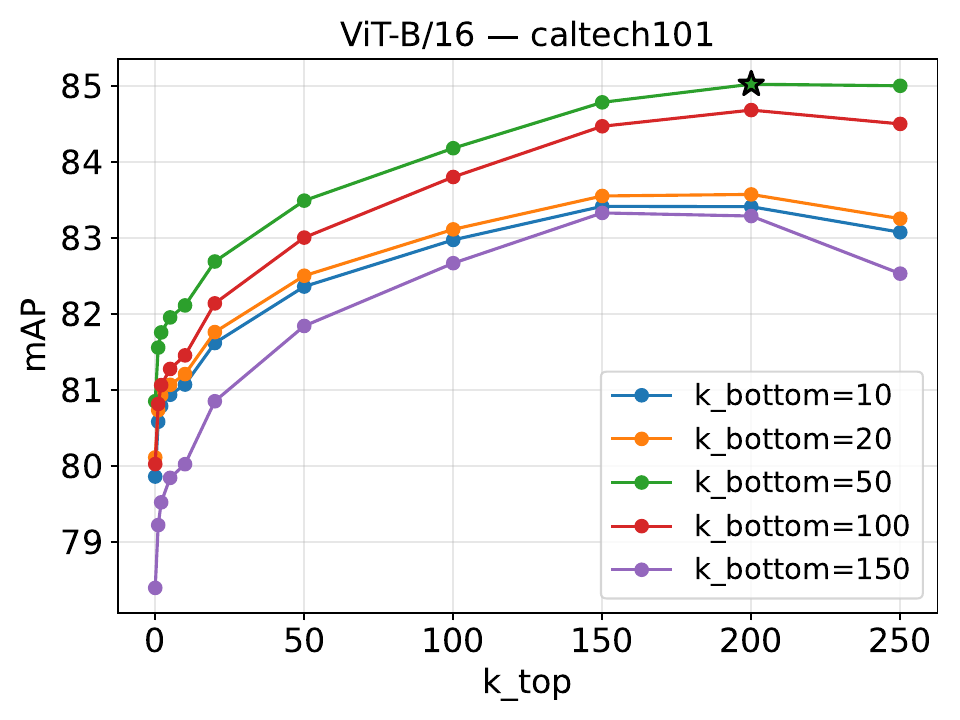}}
    \hfill
    {\includegraphics[width=0.49\textwidth]{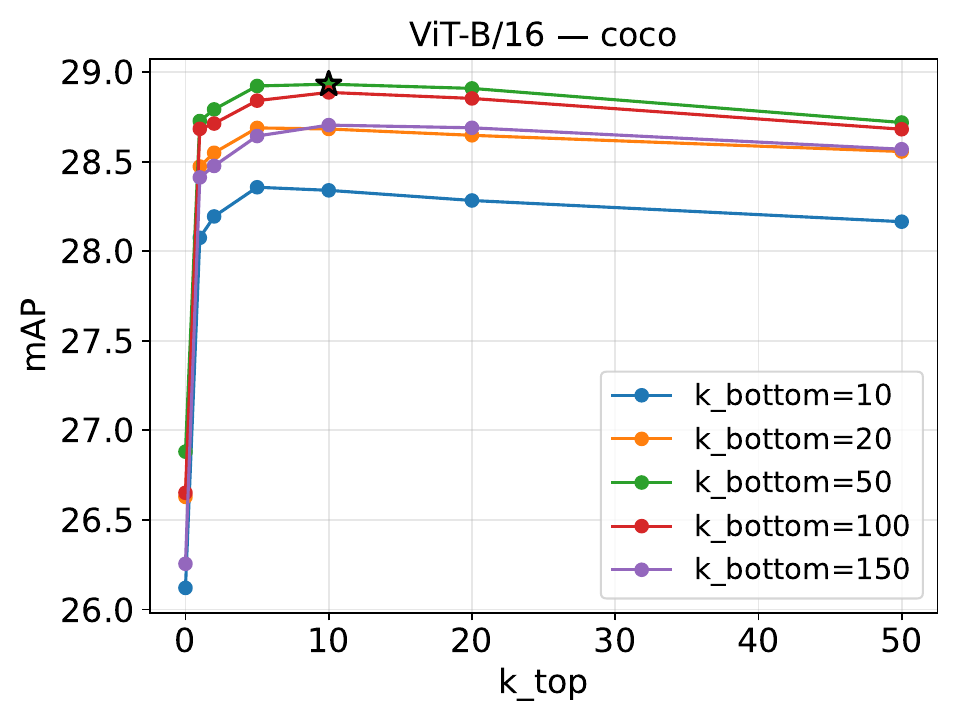}}
    \hfill
    }
    \caption{Ablation showing the impact of varying the $k_t$ and $k_b$ values for the isotropic middle band selection on Caltech101 (image-to-image retrieval) and COCO (text-to-text retrieval).}
    \label{fig:hyperparams}
\end{figure}

\minisection{Selection of $k_t$ and $k_b$.} We select $k_t$ and $k_b$, which define the middle isotropic band in IsoCLIP, for each backbone using Caltech101 (image-to-image retrieval) and COCO (text-to-text retrieval), and apply these values to all other datasets. See Sec.~\ref{sec:supp-G} of the Supplementary Material for details.

In~\cref{fig:hyperparams} we analyze the selection for ViT-B/16. On Caltech101, for $k_b=50$ the mAP improves more when varying $k_t$ from 0 to 200 (about 5\%), while for $k_t=200$, mAP improves by about 2\% when varying $k_b$. This implies that removing the top directions is more impactful for image-based tasks as shown in ~\cref{fig:method_analysis}. On COCO, we observe that the removal of only a few directions ($k_t=10, k_b=50$) is sufficient to improve the performance for text-based tasks.

\section{Conclusions}
In this paper we explored several aspects of CLIP, focusing on the role of the projectors in intra-modal misalignment. We rendered explicit an inter-modal operator hidden in CLIP responsible for aligning image and text representations, and identified a second, intra-modal operator that enforces normalization but does not promote intra-modal alignment. Based on this, we identify a well-aligned subspace characterized by the spectrum of the inter-modal operator $\Psi$, and proposed mapping projector weights onto this shared subspace to suppress anisotropic directions before embedding into the shared CLIP space. Our experiments demonstrate that, on the intra-modal tasks of image retrieval and text retrieval, IsoCLIP outperforms existing methods, induces better intra-modal alignment, and adds no latency.

\minisection{Limitations and Future Works.} Using the IsoCLIP projectors degrades performance on inter-modal tasks such as text-to-image retrieval (see Sec.~\ref{sec:supp-H} of the Supplementary Material for a discussion). Moreover, we empirically select hyperparameters $k_t$ and $k_b$ based on a single dataset for each modality, and we believe that more principled approaches for band selection should be explored in future work. 

\section*{Acknowledgments}
This work was supported by Grants PID2022-143257NB-I00, and AIA2025-163919-C52 funded by MCIN/AEI/10.13039/501100011033 and the FEDER, and by the European Union Horizon Europe research and innovation programme under grant agreement number 101214398 (ELLIOT) and the AI4Debunk project (HORIZON-CL4-2023-HUMAN-01-CNECT grant n.101135757). Bart{\l}omiej Twardowski acknowledges the grant RYC2021-032765-I and National Centre of Science (NCN, Poland) Grant No. 2023/51/D/ST6/02846.
{
    \small
    \bibliographystyle{ieeenat_fullname}
    \bibliography{main}
}

\clearpage
\onecolumn
\appendix

\setcounter{section}{0}
\setcounter{subsection}{0}
\setcounter{subsubsection}{0}
\setcounter{equation}{0}
\setcounter{figure}{5}
\setcounter{table}{4}
\renewcommand{\theHequation}{supp.\arabic{equation}}
\renewcommand{\theHfigure}{supp.\arabic{figure}}
\renewcommand{\theHtable}{supp.\arabic{table}}
\renewcommand{\theHsection}{supp.\Alph{section}}
\renewcommand{\theHsubsection}{supp.\Alph{section}.\arabic{subsection}}
\renewcommand{\theHsubsubsection}{supp.\Alph{section}.\arabic{subsection}.\arabic{subsubsection}}

\makeatletter
\let\ps@plain\ps@empty
\let\oldaddcontentsline\addcontentsline
\renewcommand{\addcontentsline}[3]{%
  \def\tempa{#2}%
  \def\secname{section}%
  \def\subsecname{subsection}%
  \def\subsubsecname{subsubsection}%
  \ifx\tempa\secname
    \oldaddcontentsline{stc}{section}{#3}%
  \else\ifx\tempa\subsecname
    \oldaddcontentsline{stc}{subsection}{#3}%
  \else\ifx\tempa\subsubsecname
    \oldaddcontentsline{stc}{subsubsection}{#3}%
  \fi\fi\fi
}
\makeatother

\renewcommand{\thesubsection}{\thesection.\arabic{subsection}}
\renewcommand{\thesubsubsection}{\thesubsection.\arabic{subsubsection}}

\renewcommand{\theequation}{S\arabic{equation}}
\renewcommand{\thetable}{S\arabic{table}}
\renewcommand{\thefigure}{S\arabic{figure}}

\renewcommand{\cftsecfont}{\bfseries}
\renewcommand{\cftsecpagefont}{\bfseries}

\makeatletter
\renewcommand{\numberline}[1]{\makebox[2.8em][c]{#1}\hspace{0.12em}}
\makeatother
 
\section*{Supplementary Material}
 
\noindent This supplementary material provides additional details and analyses that complement the main paper. We include the full gradient derivation of the CLIP loss (Sec.~A), a detailed analysis of how IsoCLIP improves intra-modal retrieval (Sec.~B), and an investigation of the effect of adding top and bottom spectral directions to the middle band (Sec.~C). We then describe the extension of IsoCLIP to non-linear projection heads (Sec.~D), followed by dataset descriptions and implementation details (Sec.~E). Additional experimental results, ablations, and comparisons are reported in Sec.~F. Finally, we discuss hyperparameter selection (Sec.~G), inter-modal degradation (Sec.~H), and provide the complete IsoCLIP algorithm pseudocode (Sec.~I).

\begingroup
\hypersetup{linkcolor=black}
\setcounter{tocdepth}{2}

\begin{center}
    \large\bfseries Table of Contents
\end{center}

\vspace{0.5em}
\makeatletter
\let\oldcontentsline\contentsline
\renewcommand{\contentsline}[4]{%
  \oldcontentsline{#1}{#2}{}{#4}%
}
\@starttoc{stc}
\let\contentsline\oldcontentsline
\makeatother
\endgroup

\clearpage
\section{Inter-Modal and Intra-Modal Operators: Gradient Derivation for the CLIP Loss}
\label{sec:supp-A}

The symmetric contrastive loss of CLIP is defined as:
\begin{equation}
    \mathcal{L}_{\text{CLIP}} = \frac{1}{2} (\mathcal{L}_{i \rightarrow t} + \mathcal{L}_{t \rightarrow i}  ), 
\end{equation}
where $\mathcal{L}_{i \rightarrow t}$ moves the embedding of each image $i$ toward the corresponding \textit{positive} paired text $t$, while pushing it away from all other texts in the mini-batch. In the main paper we present the gradient contribution of the \textit{positive} text  $g_t$; here we provide the complete derivation. 

The loss $\mathcal{L}_{i \rightarrow t}$ is defined as:
\begin{equation}
\mathcal{L}_{i\rightarrow t}
= \; - \log 
\frac{
    \exp\!\big( \text{sim}(f_i, g_t)/\tau \big)
}{
    \displaystyle
    \sum_{t'} \exp\!\big( \text{sim}(f_i, g_{t'})/\tau \big)
}
= \; - \log 
\frac{
    \exp\!\Bigg(
        \dfrac{
            f_i^{\top} (W_i^{\top} W_t) g_t / \tau
        }{
            \|W_i f_i\|_2 \, \|W_t g_t\|_2
        }
    \Bigg)
}{
    \displaystyle
    \sum_{t'}
    \exp\!\Bigg(
        \dfrac{
            f_i^{\top} (W_i^{\top} W_t) g_{t'}/ \tau
        }{
            \|W_i f_i\|_2 \, \|W_t g_{t'}\|_2
        }
    \Bigg)
},
\end{equation}
 where $f_i$ and $g_t$ denote the pre-projection image and positive text feature respectively; $W_i$ and $W_t$ denote the image and text projector weights respectively;  $\tau$ is the temperature; $t$ denotes the  \textit{positive text} for image $i$, and $t'$ ranges over the positive and negative texts in the mini-batch. 
 
Defining the normalization factor
\begin{equation}
\alpha_{t,i}
= \frac{1}{\|W_i f_i\|_2\,\|W_t g_t\|_2},
\label{eq:supp-normalization}
\end{equation}
and the logit of the positive image–text pair
\begin{equation}
\label{eq:supp-sim}
s_{t}
= \alpha_{t,i}\, f_i^{\top} (W_i^{\top} W_t) g_t.
\end{equation}
 the loss becomes:
\begin{equation*}
\mathcal{L}_{i\rightarrow t}
= - \log 
\frac{
    \exp(s_t/\tau)
}{
    \displaystyle \sum_{t'} \exp(s_{t'}/\tau)
}.
\end{equation*}
By applying the chain rule and isolating the contribution of the \textit{positive} text embedding, we obtain: 
\begin{equation}
\label{eq:supp-positive-der}
\frac{\partial \mathcal{L}_{i\rightarrow t}}{\partial f_i}
\;=\;
\frac{\partial \mathcal{L}_{i\rightarrow t}}{\partial s_{t}}
\;\frac{\partial s_{t}}{\partial f_i}.
\end{equation}
The first term is the standard derivative of  cross-entropy with respect to the logit $s_t$:
\begin{equation*}
\frac{\partial \mathcal{L}_{i\rightarrow t}}{\partial s_t}
= \frac{1}{\tau}\,\big(p_t - 1\big), \qquad \text{where} \qquad p_t
=
\frac{\exp(s_t / \tau)}{\sum_{t'} \exp(s_{t'} / \tau)}
\end{equation*}
is the softmax probability of the positive text $t$.
 
For the second term, recalling the definitions in  \cref{eq:supp-normalization} and \cref{eq:supp-sim}, we obtain
\begin{equation}
\label{eq:supp-orig_der}
\frac{\partial s_t}{\partial f_i}
= \alpha_{t,i} (W_i^{\top} W_t) g_t
   + \frac{\partial \alpha_{t,i}}{\partial f_i} \,
      f_i^{\top} (W_i^{\top} W_t) g_t = \alpha_{t,i} (W_i^{\top} W_t) g_t
   + \frac{\partial \alpha_{t,i}}{\partial f_i} \,\Big(
     \frac{s_t}{\alpha_{t,i}} \Big)
\end{equation}
Exploiting the derivative of the norm, a straightforward calculation yields
 \begin{equation}
\label{eq:supp-dalpha}
\frac{\partial \alpha_{t,i}}{\partial f_i}
=
-\frac{1}{\|W_t g_t\|_2 \|W_i f_i\|_2^{2}}\,
\frac{\partial \|W_i f_i\|_2}{\partial f_i} = 
-\alpha_{t,i} W_i^{\top} \frac{W_i f_i}{||W_i f_i ||^2_2}. 
\end{equation}
Replacing this result in \cref{eq:supp-orig_der}, gives:
\begin{equation}
\label{eq:supp-grad-f}
\frac{\partial s_{t}}{\partial f_i}
= \alpha_{t, i}\,
\overbrace{W_i^{\top}W_t}^{\Psi}\,g_{t}
- s_{t}\,
\frac{\overbrace{W_i^{\top} W_i}^{\Psi_i}\,f_i}{\|W_i f_i\|_2^{2}},
\end{equation}
where $\Psi$ and $\Psi_i$ represent respectively the \textit{inter-modal} and \textit{intra-modal} operator, matching the definition in the main paper.

\minisection{On the role of negative texts.}  
As specified in the main paper and in the previous derivation, we focus on the contribution of the positive text $g_t$ to the gradient (see \cref{eq:supp-positive-der}). In general, the full gradient with respect to $f_i$ is obtained by summing over all texts $t'$ in the batch:
\begin{equation*}
\frac{\partial \mathcal{L}_{i\rightarrow t}}{\partial f_i}
= \frac{1}{\tau}\sum_{t'} (p_{t'} - y_{t'})\,
\left[
    \alpha_{t',i} \Psi g_{t'}
    - s_{t'}\,\frac{\Psi_i f_i}{\|W_i f_i\|_2^2}
\right],
\end{equation*}
where $y_{t'} = 1$ only for the positive text. Negative texts contribute additional image–text directions $g_{t'}$, but they do not introduce new operators: all interactions between image and text features occur through the \textit{inter-modal} operator $\Psi$, while $\Psi_i$ acts solely as the \textit{intra-modal} normalization term enforcing unit-length image embeddings. Thus, negatives only reweight the contributions of these two operators: repulsion from negatives and attraction toward the positive act exclusively through the inter-modal operator $\Psi$, while $\Psi_i$ remains  a normalization term.

\section{Improving Intra-Modal Retrieval via IsoCLIP}
\label{sec:supp-B}

\begin{figure}
    \centering
\includegraphics[width=\linewidth]{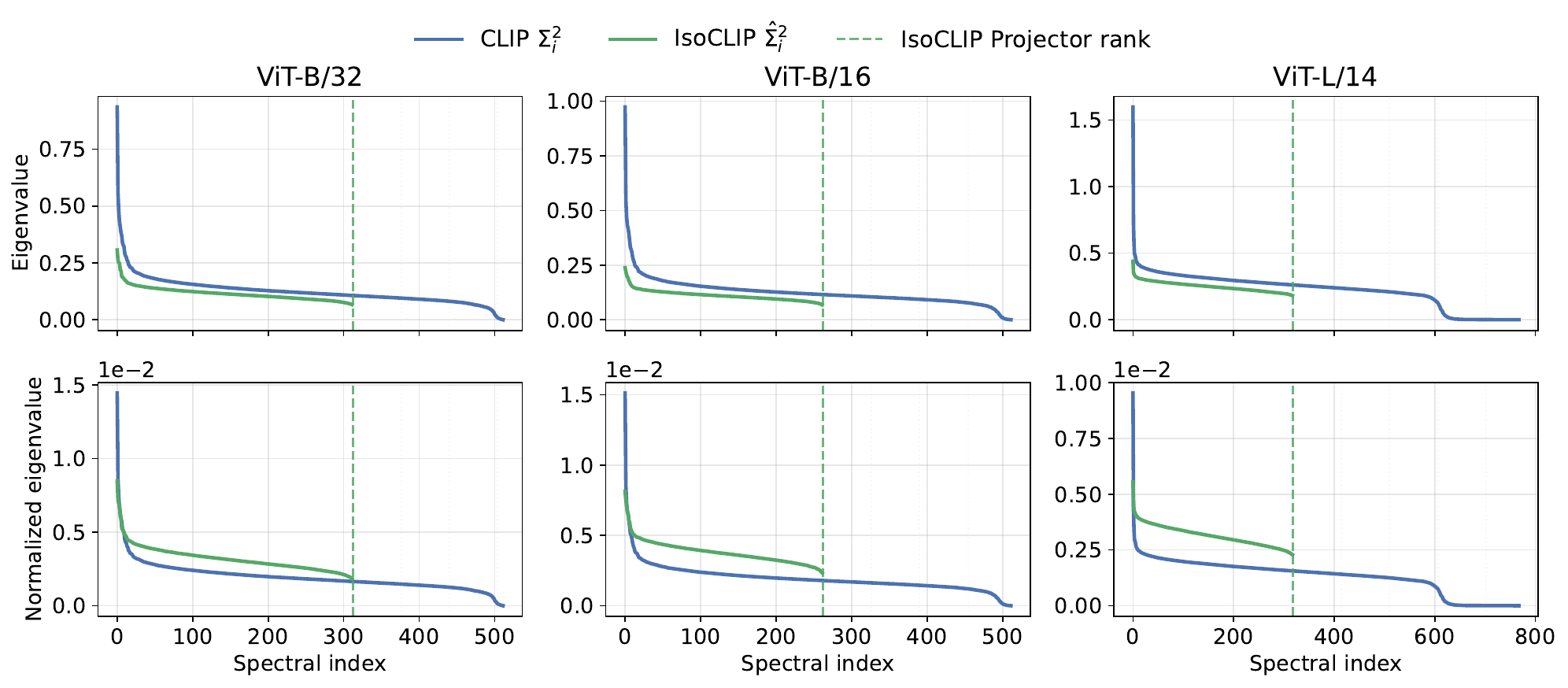}
    \caption{\textbf{Spectrum of the intra-modal operator in CLIP ($\Sigma_i^2$) and after applying IsoCLIP ($\hat{\Sigma}_i^2$)}. (Top) IsoCLIP truncates the spectrum of the intra-modal operator according to the retained subspace defined by the middle-band of the inter-modal operator (Eq.~\eqref{eq:supp-projector-weights-iso}), resulting in a lower-rank operator. (Bottom) The normalized eigenvalues (obtained by dividing each eigenvalue by the sum of the spectrum) reveal that, after applying IsoCLIP, the intra-modal operator is distributed across more directions than in standard CLIP.}
    \label{fig:supp-iso-clip-spectrum}
\end{figure}
At the end of Section \ref{sec:main-4} in the main paper we discussed that projecting the image and text projectors $W_i$ and $W_t$ onto the subspaces corresponding to the approximately isotropic region of the inter-modal operator spectrum $\Psi$ improves intra-modal retrieval performance. The projected weights are defined as:
\begin{equation}
\label{eq:supp-projector-weights-iso}
        \widehat{W}_i = W_i U_{\mathcal{S}_U} U_{\mathcal{S}_U}^{\top}, \quad     \widehat{W}_t = W_t V_{\mathcal{S}_V} V_{\mathcal{S}_V}^{\top}.
\end{equation}
 
We now provide additional insights into the mechanism responsible for this improvement. Recall that intra-modal similarity between two image features $f_i$ and $f_{\hat{i}}$ is defined as:
\begin{equation}
    \text{sim}(f_i, f_{\hat{i}})= \frac{f_{i}^{\top} (W_i^{\top} W_i) f_{\hat{i}}}{\|W_i f_i\|_2 \, \|W_i f_{\hat{i}}\|_2} \propto f_{i}^{\top} (W_i^{\top} W_i) f_{\hat{i}}.
\end{equation}

Let us consider the singular value decomposition of the image projector $W_i=U_i \Sigma_i V_i^{\top}$. Substituting into the similarity expression, we obtain:
\begin{equation}
    \text{sim}(f_i, f_{\hat{i}})\propto f_{i}^{\top} (V_i \Sigma_i U_i^{\top} U_i \Sigma_i V_i^{\top})f_{\hat{i}} = f_{i}^{\top} V_i \Sigma_i^2 V_i^{\top}f_{\hat{i}}.
\end{equation}

This shows that intra-modal similarity is governed by the intra-modal operator $\Psi_i= W_i^{\top}W_i = V_i \Sigma_i^2V_i^{\top}$. Moreover, this shows that CLIP image-to-image cosine similarity can be interpreted as a weighted summation over the singular directions $v_k$ of $W_i$, where the weights are given by the squared singular values $\sigma_k^2$. In practice, the spectrum $\Sigma_i^2$ is highly anisotropic, so that a small number of singular directions receive excessively large weight in similarity computations. Consequently, similarity scores are dominated by these few directions, reducing the separability between positive and negative pairs.

IsoCLIP mitigates this by restricting the projector to the middle band of the inter-modal spectrum (Eq.\ref{eq:supp-projector-weights-iso}). The resulting filtered projector $\widehat{W}_i$ can be written as $\widehat{W}_i = \hat{U}_i\hat{\Sigma}_i\hat{V}_i$. The resulting similarity becomes:
\begin{equation}
\label{eq:supp-iso-sim}
\mathrm{sim}(f_i,f_{\hat{i}})
\propto 
f_i^{\top}\!\hat{V}_i \hat{\Sigma}_i^2 \hat{V}_i^{\top} \! f_{\hat{i}}.
\end{equation}
Because IsoCLIP removes the highly anisotropic top and bottom spectral directions, the spectrum $\hat{\Sigma}^2$ becomes significantly flatter. As a result, similarity computations distribute weight across a larger number of directions corresponding to the middle band of the inter-modal spectrum, which encode cross-modal semantic alignment.

Fig.~\ref{fig:supp-iso-clip-spectrum} visualizes the eigenvalues of the intra-modal operator $W_i^{\top}W_i$, namely $\Sigma_i^2$ for the original CLIP projector and $\hat{\Sigma}_i^2$ for the IsoCLIP projector, for ViT-B/32, ViT-B/16 and ViT-L/14. The top row shows that IsoCLIP truncates the spectrum according to the retained subspace (Eq.~\eqref{eq:supp-projector-weights-iso}), resulting in a lower-rank operator. The bottom row shows the normalized spectra with the summation of the eigenvalues. Compared to CLIP, the retained IsoCLIP spectrum is less concentrated in few directions, reducing spectral anisotropy and distributing similarity across a larger set of directions. 

\begin{figure}
    \centering
    \includegraphics[width=0.8\linewidth]{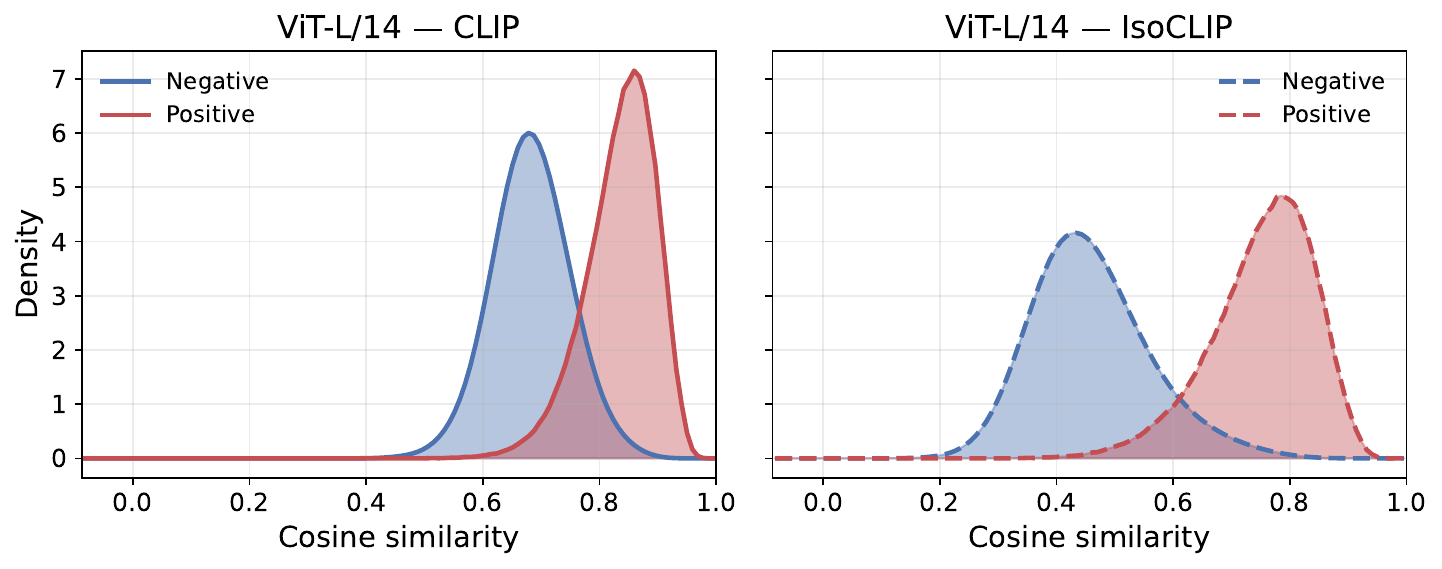}
    \caption{Cosine similarity distribution in image-to-image retrieval for positive and negative pairs on the CUB dataset for CLIP (left) and IsoCLIP (right). Positives are obtained by computing the similarity of each image with all images in the gallery that share the same label, while negatives correspond to the cosine similarity with images having different labels. We observe that the CLIP cosine similarities are concentrated in a narrower range with higher mean values, while IsoCLIP, by weighting more directions in cosine similarity computations, spreads the distribution, shifts the mean similarities to lower values and increases positive and negative separation.}
    \label{fig:supp-iso-clip-similarity-distribution}
\end{figure}

\minisection{IsoCLIP similarity distribution and mAP improvement.} It is interesting to observe the effect of the spectra $\Sigma_i^2$ and $\hat{\Sigma}_i^2$ on the cosine similarity distribution. Figure~\ref{fig:supp-iso-clip-similarity-distribution} shows the distribution of cosine similarities for positive and negative image pairs using the original CLIP projector (left) and the IsoCLIP projector (right) on CUB images. 

In Fig.~\ref{fig:supp-iso-clip-similarity-distribution}, we observe that the original CLIP projector (left), due to the strong anisotropy, concentrates both the positive and negative cosine similarity distribution in a narrow range (peaks $\approx 0.6$ for negative and $\approx 0.9$ for positive). This indicates that projected image features occupy a relatively small region of the hyper-sphere.
 
 In contrast, IsoCLIP projects $W_i$ onto the middle-band of the inter-modal operator spectrum, distributing similarity across more directions (Eq.~\ref{eq:supp-iso-sim}). As a result, cosine similarities become less concentrated and shift toward lower values (around $\approx 0.4$ for negatives and $\approx 0.8$ for positives), indicating that features occupy a larger region of the hypersphere. The reduced overlap between positive and negative similarities (shaded area) leads to higher mAP.

 \section{Adding Top and Bottom Directions to the Isotropic Middle Band Directions}
 \label{sec:supp-C}
\begin{figure} 
    \centering
    \includegraphics[width=0.5\linewidth]{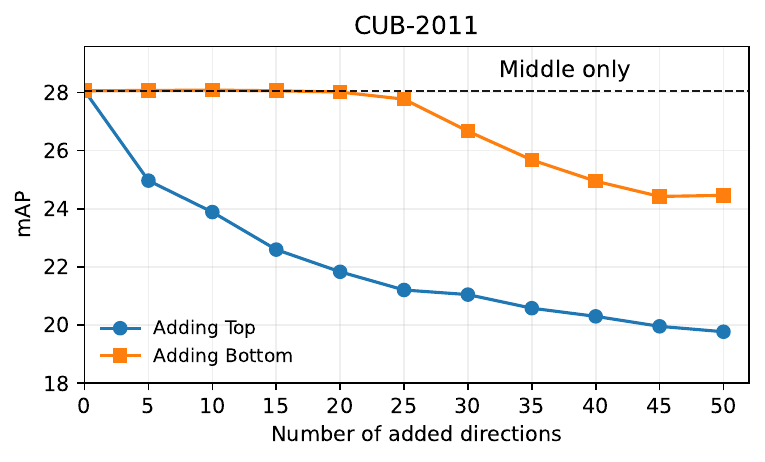}
    \caption{Image-image retrieval mAP performance obtained by iteratively adding top and bottom directions to the 50 middle-band directions when computing IsoCLIP on CUB-2011 dataset using the ViT-B/16.}
    \label{fig:supp-cub-middle-one-hand}
\end{figure}
We complement the experiments in the main paper by incrementally adding top and bottom directions to the 50 middle-band directions shown in Fig.~3 (main paper) on CUB using ViT-B/16. In Fig.~\ref{fig:supp-cub-middle-one-hand} we observe that extending the middle band with either top or bottom directions to define the subspace used to project the projector weights in IsoCLIP (Eq.~\ref{proj:equation} in the main paper) degrades mAP retrieval performance, with top directions being more detrimental than bottom ones. This result strengthens the claim that the middle region identified by the inter-modal operator is optimal for improving image-to-image retrieval performance, while the extremes, capturing modality-specific variations, are detrimental.

\section{Extension of IsoCLIP to non-linear projection heads}
\label{sec:supp-linearization}
Non-linear projection heads prevent the direct application of IsoCLIP. For instance, models like SigLIP2~\cite{tschannen2025siglip2multilingualvisionlanguage} employ a Multilayer Perceptron (MLP) head for the image encoder and a linear layer for the text encoder, mapping both modalities into the shared embedding space. We now discuss how to generalize IsoCLIP with non-linear projection heads.

Recent work~\cite{Haoqi_2024_ECCV} has shown that linear approximation of Vision Transformer blocks can reveal singular defects in attention feature maps by exploiting data-driven linearization. We propose a simple data-free first-order linearization of the final MLP layer before applying IsoCLIP in models like SigLIP2.

The last MLP visual head of SigLIP2 is defined as:
\begin{equation}
x \leftarrow x + W_2 \phi \left(W_1 \mathrm{LN}(x) + b_1\right)+ b_2,
\end{equation}
where $W_1 \in \mathbb{R}^{m \times n}, W_2 \in \mathbb{R}^{n \times m}$ are weight matrices, $b_1\in \mathbb{R}^{m}, b_2 \in \mathbb{R}^{n}$ are bias terms and  $\phi$ denotes the GELU activation. 

The Layer Normalization (LN) operator is defined as:
\begin{equation}
\mathrm{LN}(x)=\gamma \odot \frac{x - \mu(x)}{\sqrt{\sigma^2(x) + \varepsilon}} + \beta  
\end{equation}
where $\gamma$ and $\beta$ are affine parameters, $\varepsilon$ is a small constant, and $\mu(x), \sigma^2(x)$ denote the mean and the variance of $x$.
 
Writing $\mathrm{LN}(x)=\gamma\odot\hat{x}+\beta$, the affine parameters can be absorbed into the
first linear layer as $\tilde{ W}_1=W_1\operatorname{Diag}(\gamma)$ and
$\tilde{b}_1=W_1\beta+b_1$, yielding
\begin{equation*}
   x \leftarrow x + W_2\,\phi(\tilde W_1 \hat{x}+\tilde b_1)+b_2 . 
\end{equation*}
Assuming a normalized regime where LayerNorm acts approximately as identity (i.e. $\hat{x}\approx x$), and approximating GELU by its average slope
$\phi(z)\approx\tfrac{1}{2} z$, we obtain the following linearization
\begin{equation}
x \leftarrow \left(I+\frac{1}{2} W_2\tilde W_1\right)x
+ \frac{1}{2} W_2\tilde b_1 + b_2 .
\end{equation}
Accordingly, the effective image projection matrix can be written as
\begin{equation}
W_{i}=
\begin{bmatrix}
I+\tfrac12 W_2\tilde W_1 &
\tfrac12 W_2\tilde b_1 + b_2
\end{bmatrix},
\end{equation}
which can be used to apply IsoCLIP as in Eq.~\ref{eq:supp-projector-weights-iso}.

\section{Dataset and Implementation Details}
 \label{sec:supp-E}
 
In this section, we describe the datasets used and provide implementation details for the modality inversion baselines reported in the tables in the main paper.

\subsection{Datasets} For image-to-image retrieval, we consider 13 datasets - ROxford5k~\cite{radenovic2018revisiting}, RParis6k~\cite{radenovic2018revisiting}, CUB~\cite{wah2011caltech}, Stanford Cars~\cite{krause20133d}, Oxford-IIIT Pets~\cite{parkhi2012cats}, Oxford 102 Flowers~\cite{nilsback2008automated}, FGVC Aircraft~\cite{maji2013fine}, SUN397~\cite{xiao2010sun}, Caltech101~\cite{fei2004learning}, DTD ~\cite{cimpoi2014describing}, EuroSAT~\cite{helber2019eurosat}, Food101~\cite{bossard2014food}, and UCF101~\cite{soomro2012ucf101}. We follow the dataset splits proposed in~\cite{mistretta2025cross} for all datasets. For experiments on ROxford and RParis, we include the R1M distractor set, having about 1 million images, as negative samples for all queries. Here we only report results on the Easy setting from~\cite{radenovic2018revisiting}. For the other datasets, we use the training data split as the gallery and the test split as the query set for 12 datasets except CUB where the entire dataset is considered as both the query and gallery.

For text--to-text retrieval, we consider 3 image-captioning datasets - Flickr30k~\cite{plummer2015flickr30k}, COCO~\cite{lin2014microsoft}, and NoCaps~\cite{agrawal2019nocaps}. These datasets contain multiple short captions for each image. Following~\cite{mistretta2025cross}, we consider the first caption of every image as the query and all captions in the dataset as the gallery. The goal is to retrieve the other captions which are associated to the same image. We ignore the images here for the text retrieval task. On an average, COCO and Flickr30K contains 5 captions for each image and the nocaps dataset has 10 captions for each image. We use the captions from the test split for COCO and Flickr30K following the split from~\cite{karpathy2015deep}. For nocaps, we use the validation set.

We also evaluate image classification on 10 datasets drawn from those used for image retrieval. For this setting, we use the original training splits to compute the class-wise prototypes, and we report performance on the original test splits.

\subsection{Implementation Details for Modality Inversion Baselines}
 We provide implementation details for the Optimization-based Textual Inversion (OTI) and Optimization-based Visual Inversion (OVI) methods used as baselines in our experiments from~\cite{mistretta2025cross}. These methods map features from one modality to the complementary modality space through iterative optimization.  

\minisection{Optimization-based Textual Inversion (OTI).} OTI maps image features to the text embedding space by optimizing a set of pseudo-tokens $v^* = \{v_1^*, \ldots, v_R^*\}$ in the token embedding space. Following the original implementation~\cite{mistretta2025cross}, we use a single pseudo-token ($R=1$) for all backbones, randomly initialized and concatenated with the template sentence ``a photo of". We employ the AdamW optimizer with learning rate 0.02, $\beta_1 = 0.9$, $\beta_2 = 0.999$, and weight decay 0.01, performing 150 optimization steps with mixed precision training.  

\minisection{Optimization-based Visual Inversion (OVI). } OVI maps text features into the image embedding space by optimizing a set of visual pseudo-patches $w^* = \{w_1^*, \ldots, w_P^*\}$ in the patch embedding space. Following the original implementation~\cite{mistretta2025cross}, we adopt the same optimizer settings as OTI, but we run 1000 optimization steps.

Unlike OTI, the optimal number of pseudo-patches $P$ for OVI depends on the model architecture. Following the procedure described in \cite{mistretta2025cross}, we validated the pseudo patches $P \in \{1, 2, 4, 8, 16\}$  on text-to-text retrieval on Flickr30k validation set, for both ViT-B/16 and ViT-B/16-open, which we introduce in this paper. Table~\ref{tab:supp-ovi-ablation-new} reports the results of this ablation study. For OpenAI ViT-B/16, the best performance is obtained with $P=4$ pseudo-patches (53.1\% mAP). For OpenCLIP ViT-B/16, a single pseudo-patch ($P=1$) is sufficient, achieving 60.2\% mAP.

\begin{table}
\centering
\caption{Ablation study on the number of OVI pseudo-patches $P$ for text-to-text retrieval on Flickr30K validation set. The highest mAP score for each model is highlighted in bold, with the corresponding value of $P$ used in all experiments.}
\label{tab:supp-ovi-ablation-new}
\resizebox{0.4\columnwidth}{!}{
\begin{tabular}{lc ccccc}
\toprule
& \multicolumn{5}{c}{Number of Pseudo-Patches $P$} \\ \cmidrule(lr){2-6}
Backbone & 1 & 2 & 4 & 8 & 16 \\
\midrule
ViT-B/16 & 52.8 & 52.9 & \textbf{53.1} & 51.9 & 50.8 \\
ViT-B/16-open & \textbf{60.2} & 59.1 & 58.0 & 57.3 & 57.2 \\
\bottomrule
\end{tabular}}
\end{table}

\section{Additional Results and Ablations}
 \label{sec:supp-F}
 
In this section, we provide additional results and ablations that complement those in the main paper.

\subsection{Image-to-Image Retrieval on ViT-B/32-open, EVA-02 B/16 and SigLIP2 B/16}
\begin{table*}[h]
\centering
\setlength{\tabcolsep}{3pt}
\renewcommand{\arraystretch}{1.05}
\caption{\textbf{Image-to-image retrieval} performance using \textit{OpenCLIP} ViT-B/32, pre-trained on DataComp dataset, EVA-02 B/16, pretrained on Merged-2B~\citep{FANG2024105171} and SigLIP2 B/16 pre-trained on  WebLI ~\cite{tschannen2025siglip2multilingualvisionlanguage}.}
\label{tab:supp-iso_comparison_b32-open}
\resizebox{\textwidth}{!}{
\begin{tabular}{lccccccccccccccccc}
\toprule
\textbf{Method} & \textbf{Intra-modal} &\textbf{Backbone}  & \textbf{Caltech} & \textbf{CUB} & \textbf{ROxford} & \textbf{RParis} & \textbf{Cars} & \textbf{Pets} & \textbf{Flowers} & \textbf{Aircraft} & \textbf{DTD} & \textbf{EuroSAT} & \textbf{Food101} & \textbf{SUN397}  & \textbf{UCF101} & \textbf{Avg} \\
\midrule

Image-Image & \Checkmark & \multirow{3}{*}{ViT-B/32-open}  & 82.3 & 32.1 & 50.8 & 74.7 & 46.7 & 44.1 & 77.0 & 19.6 & 36.9 & \underline{56.4} & 39.6 & 36.2 & 45.7 & 49.4  \\
OTI (I$\rightarrow$T) &  \XSolidBrush & &   \underline{83.3} & \textbf{34.3} & \underline{54.4} & \textbf{75.8} & \textbf{50.5} & \textbf{50.5} & \underline{78.0} & \textbf{20.1} & \textbf{40.9} & 54.5 & \textbf{42.9} & \textbf{37.8} & \textbf{48.2} & \textbf{51.6}\\
\hlrow IsoCLIP & \Checkmark  &  & \textbf{83.4} & \underline{34.2} & \textbf{56.8} & \textbf{75.8} & \underline{49.9} & \underline{47.8} & \textbf{78.2} & \underline{19.8} & \underline{37.6} & \textbf{57.7} & \underline{41.5} & \underline{36.6} & \underline{46.3} & \underline{51.2}  \\
 \midrule
Image-Image & \Checkmark & 
& 86.9 & 55.9 & 52.1 & 78.3  & 49.4 & 55.4  & 91.5 & 24.9 &  35.8 &   61.2 & 55.5  & 41.1 & 57.0 & 57.3   \\
\hlrow IsoCLIP & \Checkmark  & \multirow{-2}{*}{EVA-02 B/16}
& \textbf{89.7}  & \textbf{58.3}  & \textbf{53.0}  & \textbf{80.4}  & \textbf{55.2}  & \textbf{62.8} &  \textbf{92.7} & \textbf{25.7}  & \textbf{40.0}  & \textbf{62.3} & \textbf{57.5}  & \textbf{42.2}  & \textbf{57.9} & \textbf{59.8}   \\
\midrule
Image-Image & \Checkmark  & 
& 89.2 & 38.1 & 53.2 & 76.5 & 70.8 & 56.6 & \textbf{89.3} & \textbf{41.8} & 39.0 & 50.8 & 59.2 & 43.6 & 59.2  & 59.0  \\[2.4pt]
\hlrow IsoCLIP &  \Checkmark  & \multirow{-2}{*}{SigLIP2 B/16}
& \textbf{93.1} & \textbf{41.4} & \textbf{54.2} & \textbf{77.9} & \textbf{74.5} & \textbf{64.0} & 87.2 & 40.9 & \textbf{43.8} & \textbf{53.2} & \textbf{63.6} & \textbf{46.9} & \textbf{61.8} & \textbf{61.7}  \\
 
\bottomrule
\end{tabular}
}
\end{table*}

In \cref{tab:supp-iso_comparison_b32-open}, we compare IsoCLIP against standard image-to-image retrieval (Image-Image) using only the vision encoder ViT-B/32-open pretrained on the DataComp dataset, as well as against the textual inversion–based approach (OTI). Consistent with the results in the main paper, where ViT-B/16-open with the same pre-training was used, IsoCLIP performs significantly better than Image-Image across all datasets, and achieves slightly lower performance than OTI while requiring substantially lower query latency. 

We also evaluate IsoCLIP on EVA-02 B/16 pre-trained on the Merged-2B dataset, where it consistently outperforms standard image-to-image retrieval. Finally, we evaluate SigLIP2 B/16 pre-trained on WebLI, and observe that, despite the linearization of the last MLP projection head (Sec.~\ref{sec:supp-linearization}), IsoCLIP significantly improves performance on most datasets.

\subsection{Image Classification on ViT-B/16-open}
\begin{table*}[h]
\centering
\setlength{\tabcolsep}{3pt}
\renewcommand{\arraystretch}{1.05}
\caption{\textbf{Image classification} performance using OpenCLIP ViT-B/16, pre-trained on DataComp dataset, on 10 datasets. We compare IsoCLIP  with intra-modal NCM classification and zero-shot classification.}
\label{tab:supp-iso_comparison_classification}
\resizebox{\textwidth}{!}{
\begin{tabular}{lcccccccccccccc}
\toprule
\textbf{Method} & \textbf{Intra-modal} & \textbf{Classifier} & \textbf{Backbone} & \textbf{Caltech} & \textbf{Cars} & \textbf{Pets} & \textbf{Flowers} & \textbf{Aircraft} & \textbf{DTD} & \textbf{EuroSAT} & \textbf{Food101} & \textbf{SUN397}  & \textbf{UCF101} & \textbf{Average} \\
\toprule
  
Image-Text & \XSolidBrush & Zero-Shot  & \multirow{3}{*}{ViT-B/16-open} & \underline{96.9} & 89.8 & \textbf{92.8} & 75.3 & 29.8 & 58.3 & 53.4 & \textbf{87.5} & 69.8 & 67.8 & 72.1 \\
Image-Image & \Checkmark & NCM &  & 96.7 & \underline{90.6} & 88.9 & \underline{98.6} & \underline{54.0} & \underline{74.5} & \underline{84.4} & 86.7 & \textbf{74.9} & \underline{79.9} & \underline{82.9} \\
\hlrow IsoCLIP& \Checkmark & NCM  &  & \textbf{97.2} & \textbf{91.4} & \underline{90.6} & \textbf{98.8} & \textbf{54.5} & \textbf{75.9} & \textbf{85.5} & \underline{87.1} & \underline{74.3} & \textbf{80.9} & \textbf{83.6} \\
\bottomrule
\end{tabular}
}
\end{table*}
We extend the experiments on image prototype-based classification with the Nearest Class Mean (NCM) classifier, presented in the main paper, by also evaluating IsoCLIP on ViT-B/16-open. For this model, we observe that IsoCLIP slightly outperforms standard NCM on the image encoder on average and provides consistent improvements on most datasets.

\subsection{Comparison with Unimodal DINOv2 B/14}

We performed a preliminary comparison with DINOv2 B/14 for image-to-image retrieval. The results are highly dataset dependent: DINOv2 achieves 67.0 mAP on CUB (vs. 53.0 for IsoCLIP PE-Core-B/16), but only  22.3 mAP on Cars, far below IsoCLIP PE-Core-B/16 (62.3 mAP), despite ViT-B/14 processing more image patches due to its smaller patch size. These results suggest that CLIP-style models remain competitive for image-to-image retrieval and highlight the utility of IsoCLIP. A more systematic comparison with self-supervised models is left for future work.

\subsection{Experiments on Places365 and iNaturalist}

\begin{wraptable}{r}{0.4\textwidth}
\vspace{-28pt}
\centering
\caption{Experiments on Places365 \& iNaturalist.}
\label{tab:supp-inat}
\resizebox{\linewidth}{!}{
\begin{tabular}{llccc}
\toprule
\textbf{Method} & \textbf{Backbone} & \textbf{Places365} & \textbf{iNat} & \textbf{Avg} \\
\midrule
Image-Image &  & 16.72 & 9.61 & 13.17 \\
\hlrow IsoCLIP & \multirow{-2}{*}{PE-Core B/16} & \textbf{17.04} & \textbf{13.07} & \textbf{15.06} \\
\midrule
Image-Image &  & 16.17 & 7.56 & 11.87 \\
\hlrow IsoCLIP & \multirow{-2}{*}{SigLIP2 B/16} & \textbf{17.78} & \textbf{8.43} & \textbf{13.11} \\
\midrule
Image-Image &  & 14.54 & 10.03 & 12.29 \\
\hlrow IsoCLIP &  \multirow{-2}{*}{EVA-02 B/16} & \textbf{15.01} & \textbf{11.23} & \textbf{13.12} \\
\bottomrule
\end{tabular}
}
\end{wraptable}
We evaluate IsoCLIP on more benchmarks like Places365~\cite{zhou2017places} and iNaturalist~\cite{van2018inaturalist} and present the results in~\cref{tab:supp-inat} using PE-Core B/16, SigLIP2 B/16 and EVA-02 B/16. We consider the validation set of Places365 having 35k images and iNaturalist 2021 mini-train version having 500k images for both gallery and query images. The values of $k_t$ and $k_b$ are selected using Caltech101. We show that IsoCLIP consistently improves accuracy on both datasets across different models. On average, IsoCLIP improves by 1.89\% on PE-Core B/16, 1.24\% on SigLIP2 B/16 and 0.83\% on EVA02 B/16.

\subsection{Additional Ablations} 
\begin{table*}[h]
\centering
\setlength{\tabcolsep}{3pt}
\renewcommand{\arraystretch}{1.05}
\caption{We compare IsoCLIP against using pre-projection image features for retrieval (Image–Image [Pre]) and against whitening the CLIP image projection weights ($W_i^{\text{white}}$), using ViT-B32 OpenAI and Open ViT-B32 and ViT-B16 pre-trained on Datacomp dataset. } 
\label{tab:supp-ablation_studies}
\resizebox{0.99\textwidth}{!}{
\begin{tabular}{llcccccccccccccc}
\toprule
\textbf{Method} & \textbf{Backbone} & \textbf{Caltech} & \textbf{CUB} &  \textbf{ROxford} & \textbf{RParis} & \textbf{Cars} & \textbf{Pets} & \textbf{Flowers} & \textbf{Aircraft} & \textbf{DTD} & \textbf{EuroSAT} & \textbf{Food101} & \textbf{SUN397} & \textbf{UCF101} & \textbf{Average} \\
\toprule
 
Image-Image & \multirow{4}{*}{ViT-B/32}  
& 77.1 & 22.9 & 42.6 & 67.9 & 24.6 & 30.5 & 62.0 & \underline{14.5} & 28.1 & 47.9 & 32.3 & 34.3 & 47.1 & 40.9 \\
Image-Image [Pre] & & 78.3 & 23.3 & 44.9 & 70.5 & 24.5 & 32.7 & 63.4 & 14.4 & \underline{29.5} & \underline{50.8} & \underline{34.2} & \underline{36.0} & \textbf{49.7} & 42.5 \\
$W_i^{\text{white}}$ &  
& \underline{78.4} & \underline{24.5} & \underline{45.0} & \underline{70.7} & \underline{25.9} & \underline{33.2} & \underline{63.7} & 14.3 & 29.4 & 50.3 & \underline{34.2} & 35.1 & \underline{48.9} & \underline{42.6} \\
\hlrow IsoCLIP &  
& \textbf{80.8} & \textbf{27.0} & \textbf{47.2} & \textbf{73.8} & \textbf{30.0} & \textbf{40.8} & \textbf{66.5} & \textbf{14.9} &\textbf{30.9} & \textbf{51.5} & \textbf{38.0} & \textbf{36.4} & 48.4 & \textbf{45.1}  \\
\midrule

Image-Image & \multirow{4}{*}{ViT-B/32-open} 
& 82.3 & 32.1 & 50.8 & \underline{74.7} & \underline{46.7} & 44.1 & \underline{77.0} & \underline{19.6} & 36.9 & 56.4 & 39.6 & 36.2 & 45.7 & 49.4 \\
Image-Image [Pre] & 
& \textbf{83.6} & 31.6 & \underline{52.8} & 73.9 & 46.1 & 44.6 & 76.5 & 19.5 & \underline{37.3} & \underline{57.5} & \underline{39.9} & \textbf{36.9} & \textbf{46.9} & \underline{49.8} \\
$W_i^{\text{white}}$&  
& 83.2 & \underline{32.9} & 52.2 & 74.1 & 45.6 & \underline{45.7} & \underline{77.0} & 18.7 & \underline{37.3} & \textbf{57.7} & 39.8 & 35.4 & 46.0 & 49.7  \\

\hlrow IsoCLIP & & \underline{83.4} & \textbf{34.2} & \textbf{56.8} & \textbf{75.8} & \textbf{49.9} & \textbf{47.8} & \textbf{78.2} & \textbf{19.8} & \textbf{37.6} & \textbf{57.7} & \textbf{41.5} & \underline{36.6} & \underline{46.3} & \textbf{51.2} \\
\midrule

Image-Image & \multirow{4}{*}{ViT-B/16-open} 
& 85.7 & 42.8 & 65.3 & 83.2 & 55.8 & 50.4 & \underline{84.6} & 23.1 & 39.9 & 57.8 & 51.1 & 39.5 &  52.9 & 56.3 \\
Image-Image [Pre] &  
& 86.3 & 42.2 & \underline{66.9} & 83.1 & 56.1 & 52.4 & 84.0 & \underline{23.2} & 39.9 & 58.0 & 52.4 & \textbf{40.9} & \textbf{54.1} & \underline{56.9} \\
$W_i^{\text{white}}$ &  
& \underline{86.4} & \underline{43.8} & 64.5 & \underline{83.5} & \underline{56.2} & \underline{54.0} & 84.4 & 22.4 & \underline{40.2} & \underline{58.1} & \underline{52.8} & \underline{39.8} & 53.3 & \underline{56.9} \\
\hlrow IsoCLIP &  
& \textbf{87.6} & \textbf{45.9} & \textbf{67.3} & \textbf{85.0} & \textbf{60.7} & \textbf{57.8} & \textbf{85.8} & \textbf{23.5} & \textbf{42.5} & \textbf{58.6} & \textbf{54.7} & 39.3 & \underline{53.4} & \textbf{58.6}  \\
\bottomrule
\end{tabular}}
\end{table*}

We provide additional ablations complementing those in the main paper, using ViT-B/32 OpenAI, Open ViT-B/32, and Open ViT-B/16 models pre-trained on the DataComp dataset. As in the main paper, we compare IsoCLIP against using the \textit{raw pre-projection features} $f_i$ (Image-Image [Pre]) and against applying whitening to the image projector weights ($W_i^{\text{white}}$). 

In \cref{tab:supp-ablation_studies}, we observe that, in general, IsoCLIP outperforms all other approaches across most datasets. Consistent with the results reported in the main paper, IsoCLIP achieves larger gains than both alternatives on the majority of benchmarks. However, when using the models pre-trained on DataComp (ViT-B/32-open and ViT-B/16-open), IsoCLIP attains performance comparable to Image-Image and other baselines on SUN397 and UCF101.

\section{Analysis of Hyperparameter Selection}
 \label{sec:supp-G}
 
All results reported in Tab.~\ref{tab:iso_comparison_imagetoimage} and Tab.~\ref{tab:iso_text_comparison} of the main paper use the same procedure for selecting the top $k_t$ and bottom $k_b$ singular directions  that define the middle-spectrum band used to align the projectors.

\begin{table} 
\centering
\caption{Selection of $k_t$ and $k_b$ for image-to-image and text-to-text retrieval using IsoCLIP. For image classification tasks, we use the same values selected for image-to-image retrieval.}
\label{tab:supp-hyperparams_all}
\resizebox{0.36\columnwidth}{!}{
\begin{tabular}{lccc}
\toprule
\textbf{Task} & \textbf{Backbone} & \textbf{$k_t$} & \textbf{$k_b$}\\
\midrule
\multirow{5}{*}{Image-Image} & ViT-B/32 & 150 & 50 \\
& ViT-B/32-open & 50 & 50 \\
& ViT-B/16 & 200 & 50 \\
& ViT-L/14 & 250 & 200 \\
& ViT-B/16-open & 100 & 100 \\
& EVA-02 B/16 & 150 & 0 \\
& PE-Core-B-16 & 300 & 50 \\
& SigLIP2 B/16 & 350 & 50 \\
\midrule
\multirow{4}{*}{Text-Text} & ViT-B/32 & 20 & 100 \\
& ViT-B/16 & 10 & 50 \\
& ViT-L/14 & 10 & 300 \\
& ViT-B/16-open & 2 & 150 \\
\bottomrule
\end{tabular}
}
\end{table}

\minisection{Selection of $k_t$ and $k_b$. }As described in the main paper, we select $(k_t, k_b)$ using \textit{Caltech101}, a generic object recognition dataset, for image-to-image retrieval and \textit{COCO} for text-to-text retrieval, and apply these values to \textit{all} the backbones.  Table~\ref{tab:supp-hyperparams_all} reports the selected values for each model. These hyperparameters differ across backbones because the shape and dimensionality of the singular spectrum of the inter-modal operator depend on the embedding dimensionality and the pre-training dataset, consistent with our analysis of the isotropic middle region in Fig.~\ref{fig:singular-values} of the main paper.

For ViT-B/16, the selected values for image-to-image retrieval are $k_t = 200$ and $k_b = 50$, validated on Caltech101 (see Figure~\ref{fig:hyperparams} left in the main paper).

\begin{figure*}[h]
    \centering
    \resizebox{\columnwidth}{!}{
    {\includegraphics[width=0.32\textwidth]{figures/b16_text_to_text/coco_text.pdf}}
    \hfill
    {\includegraphics[width=0.32\textwidth]{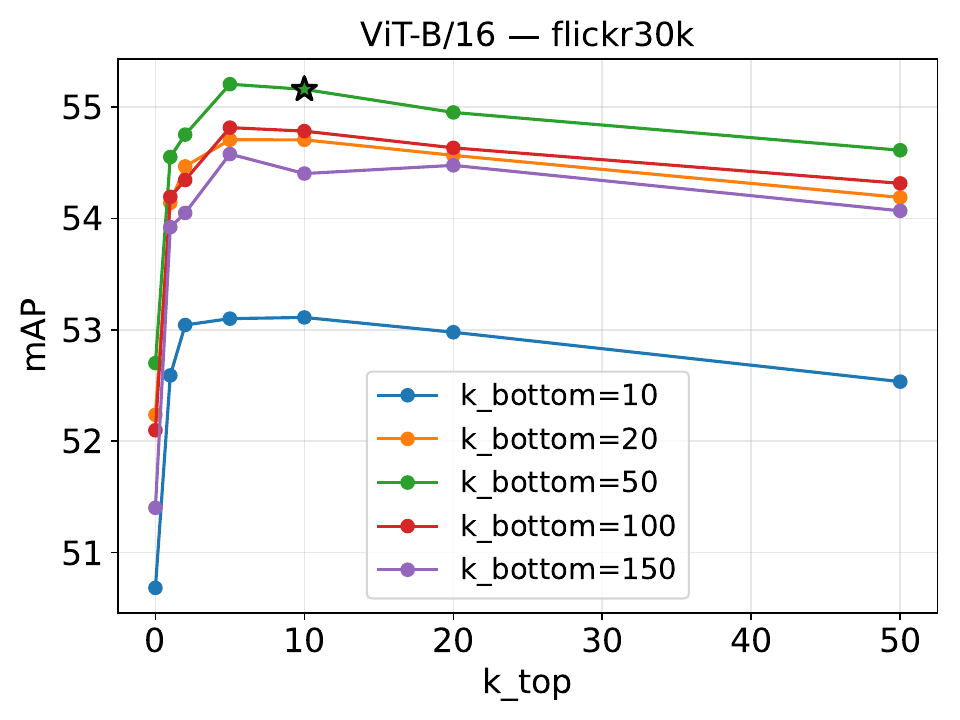}}
    \hfill
    {\includegraphics[width=0.32\textwidth]{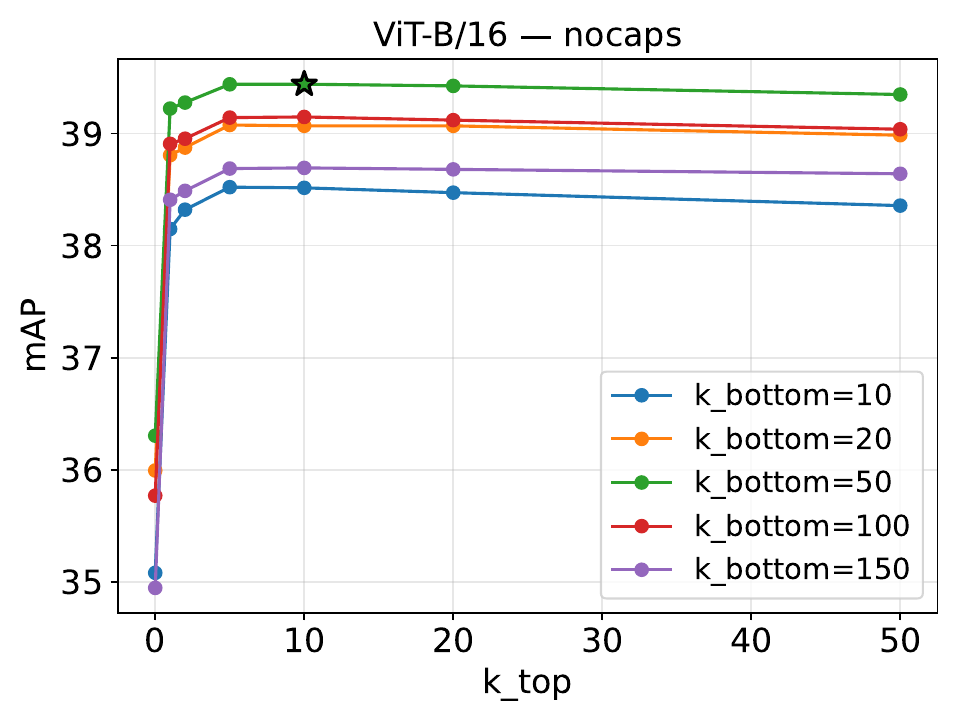}}
    }
    \caption{Analysis showing the impact of varying $k_t$ and $k_b$ for the middle band selection across datasets for text-to-text retrieval. The values used in our reported results based on selection from COCO are denoted with stars.}
    \label{fig:supp-hyperparams_text}
\end{figure*}

\begin{figure*}[t]
    \centering

    \includegraphics[width=0.32\textwidth]{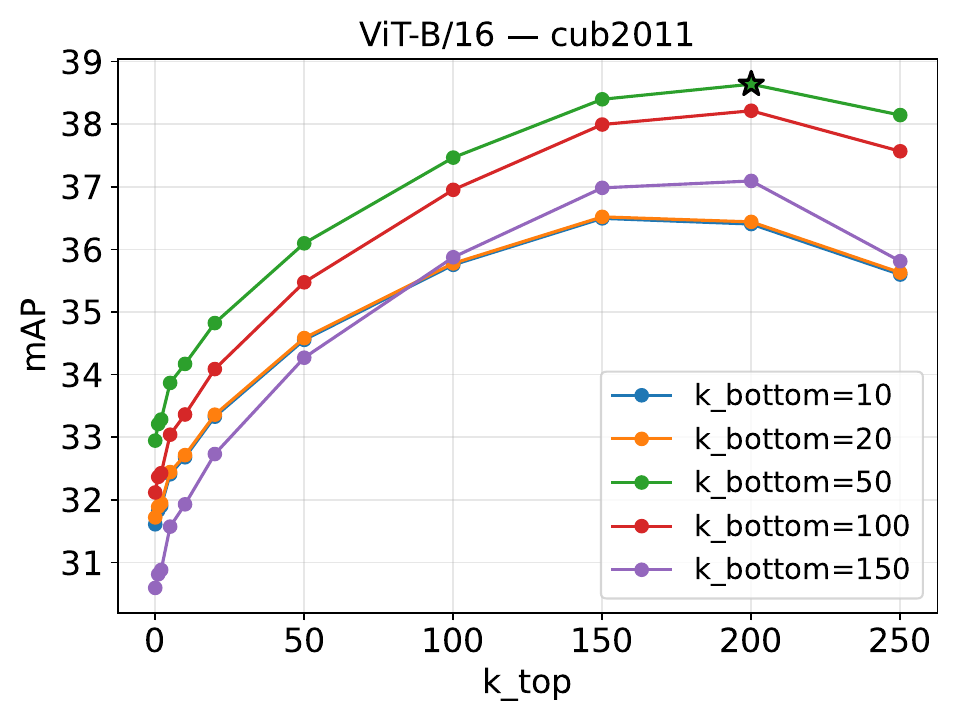}
    \includegraphics[width=0.32\textwidth]{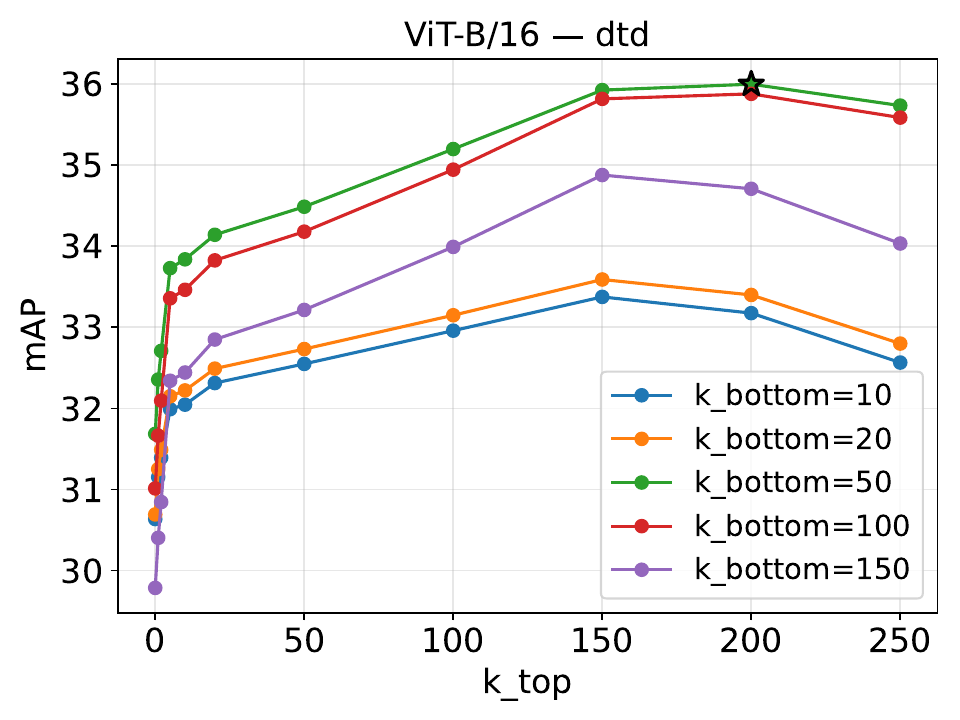}
    \includegraphics[width=0.32\textwidth]{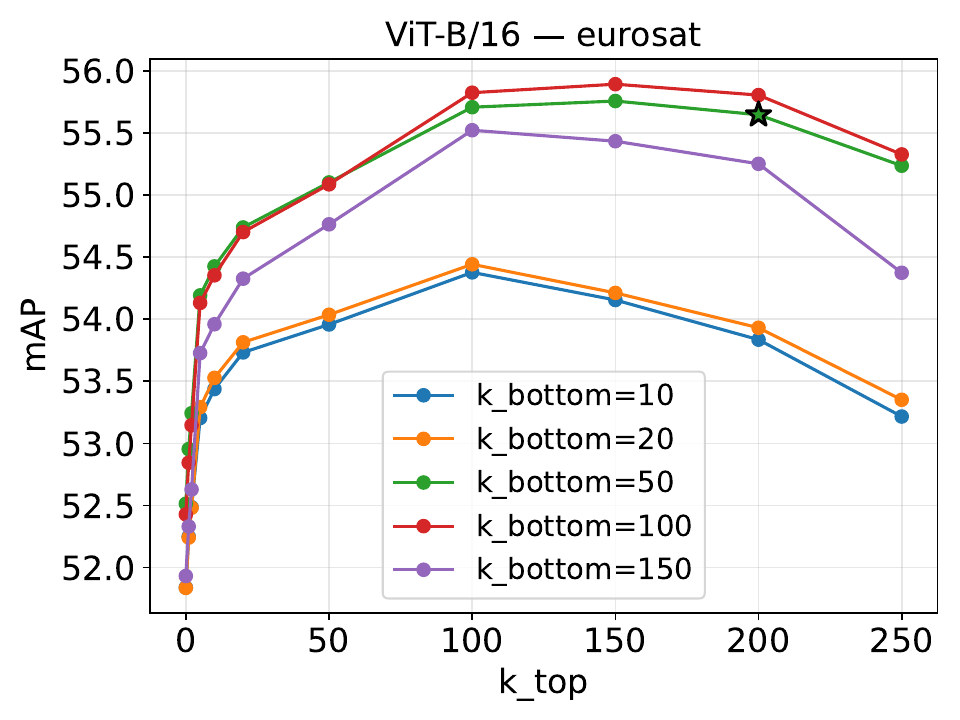}

    \includegraphics[width=0.32\textwidth]{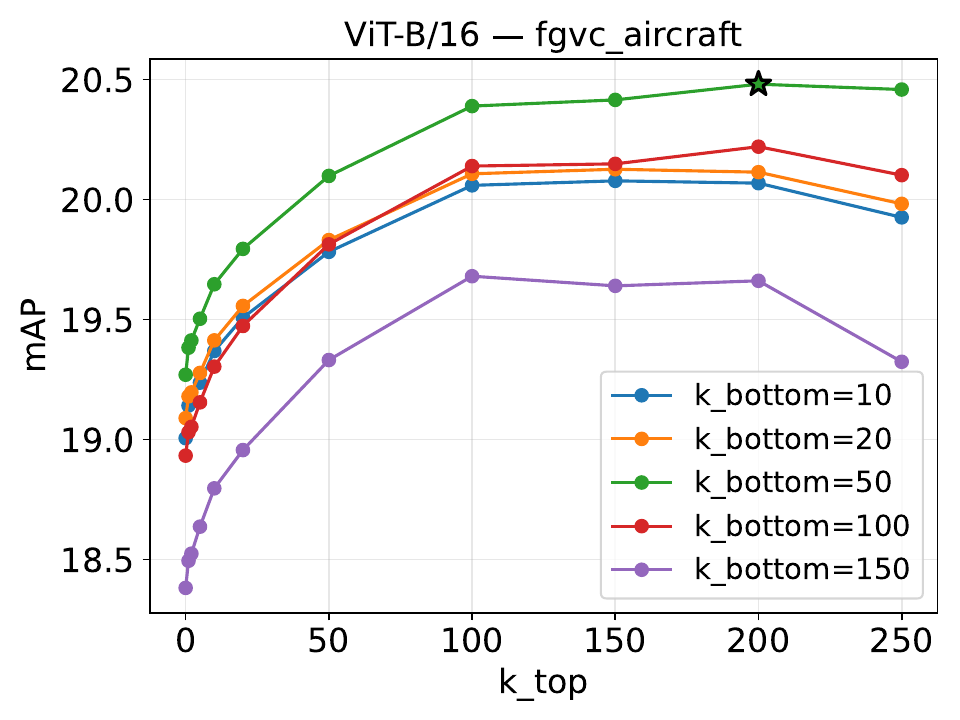}
    \includegraphics[width=0.32\textwidth]{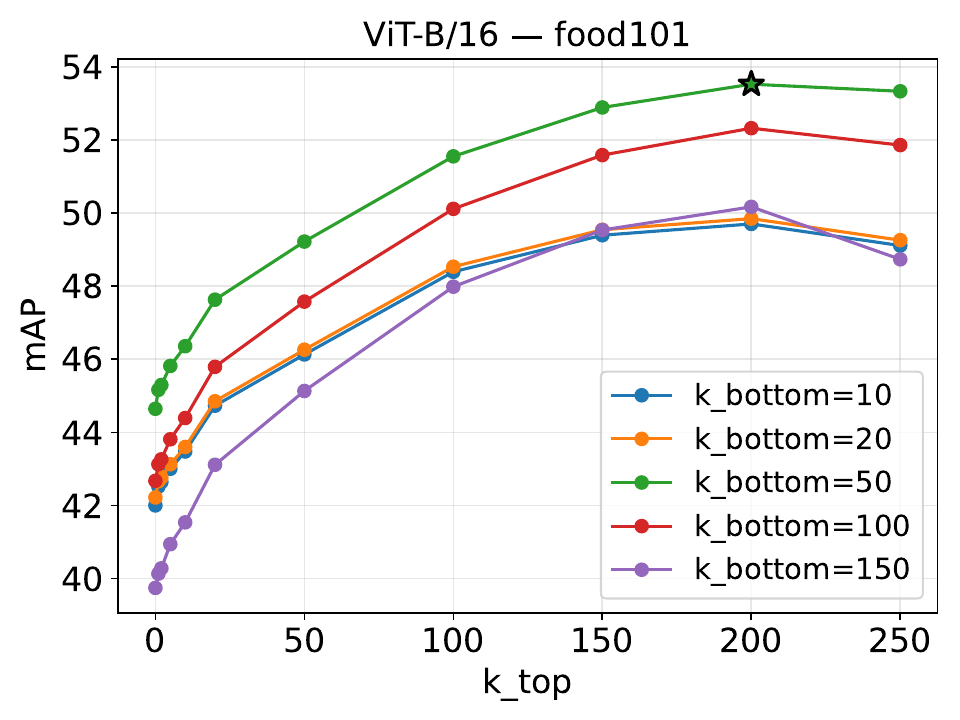}
    \includegraphics[width=0.32\textwidth]{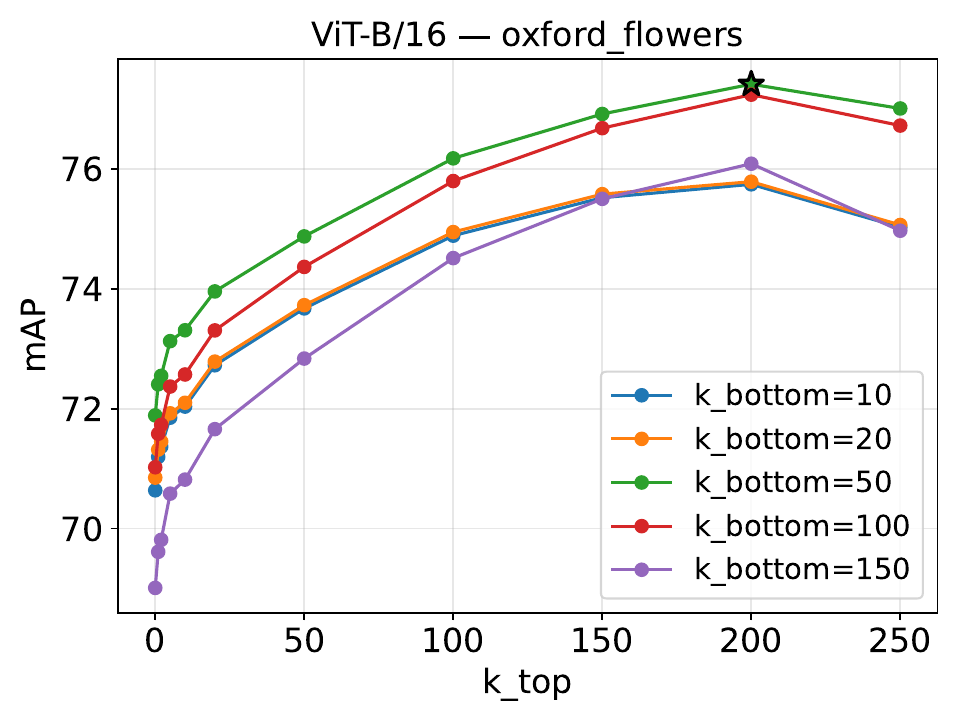}

    \includegraphics[width=0.32\textwidth]{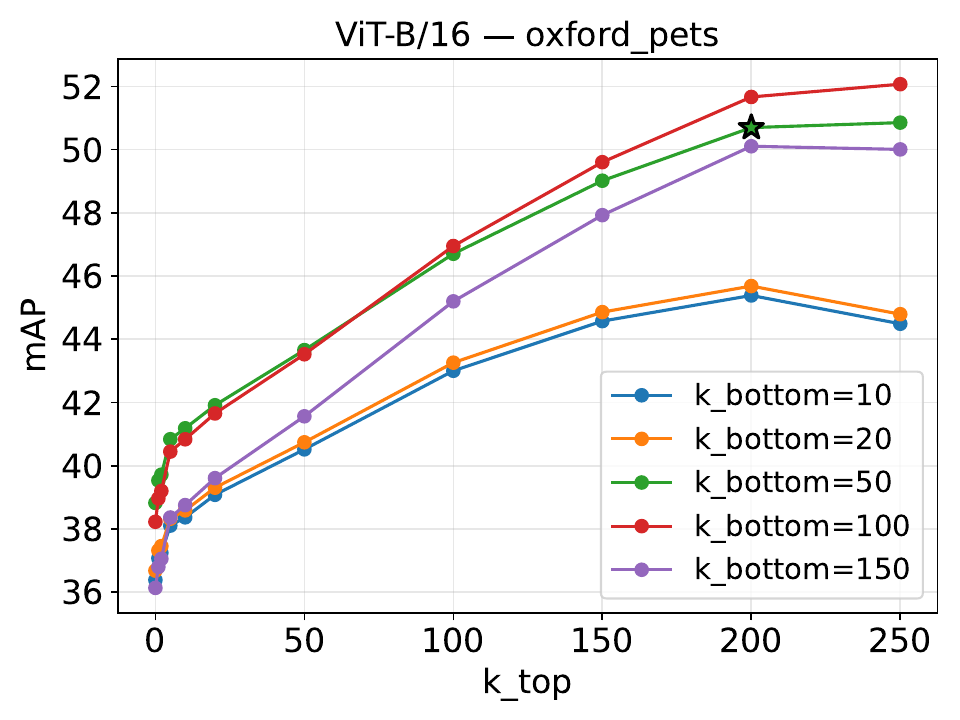}
    \includegraphics[width=0.32\textwidth]{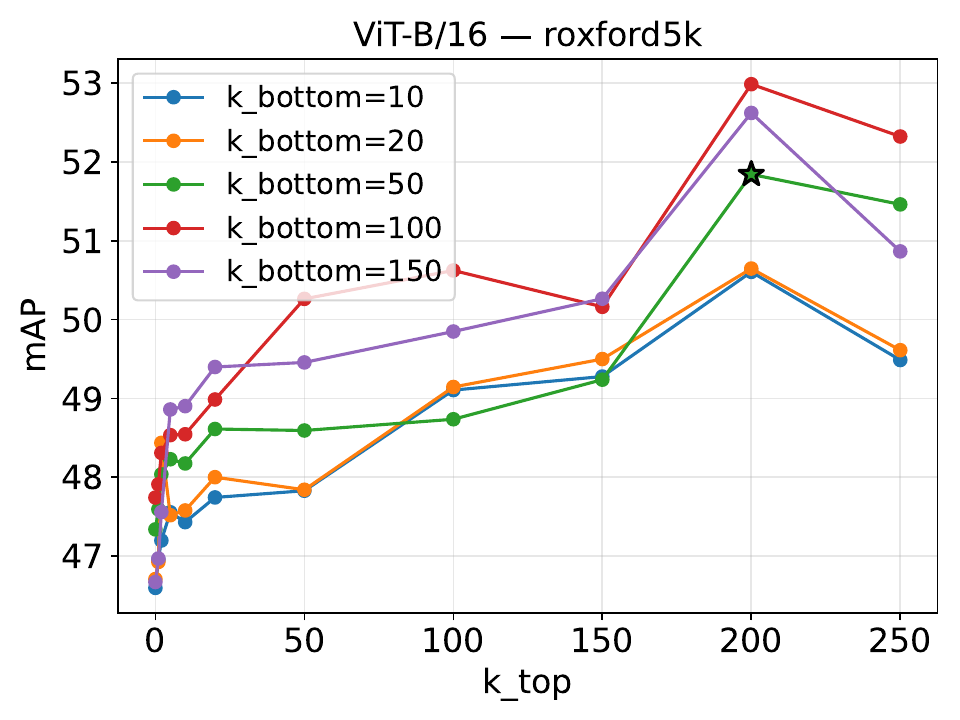}
    \includegraphics[width=0.32\textwidth]{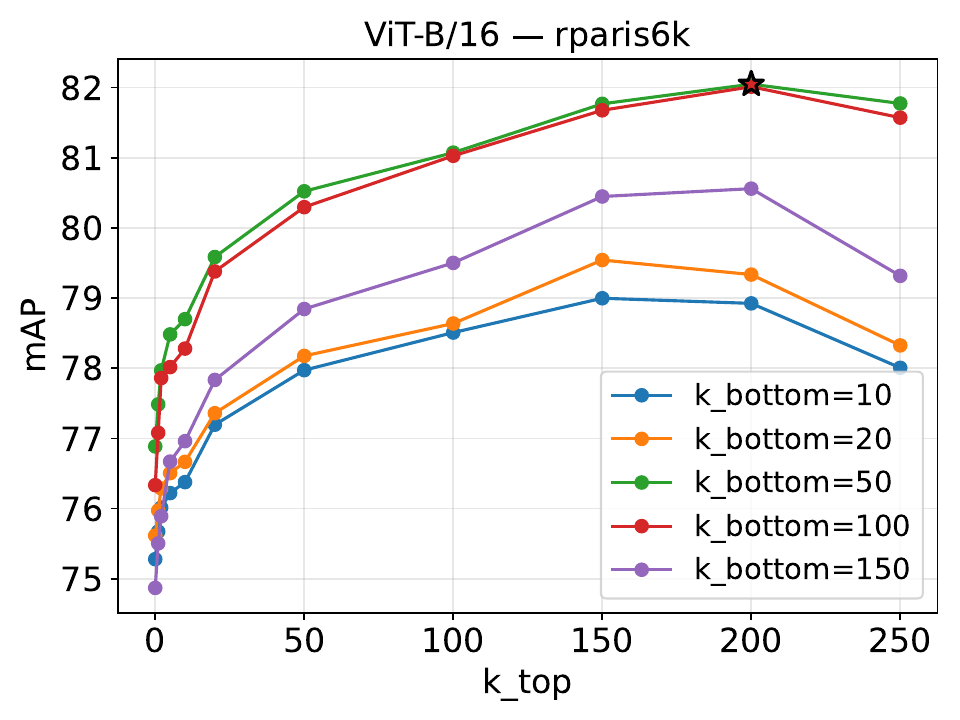}

    \includegraphics[width=0.32\textwidth]{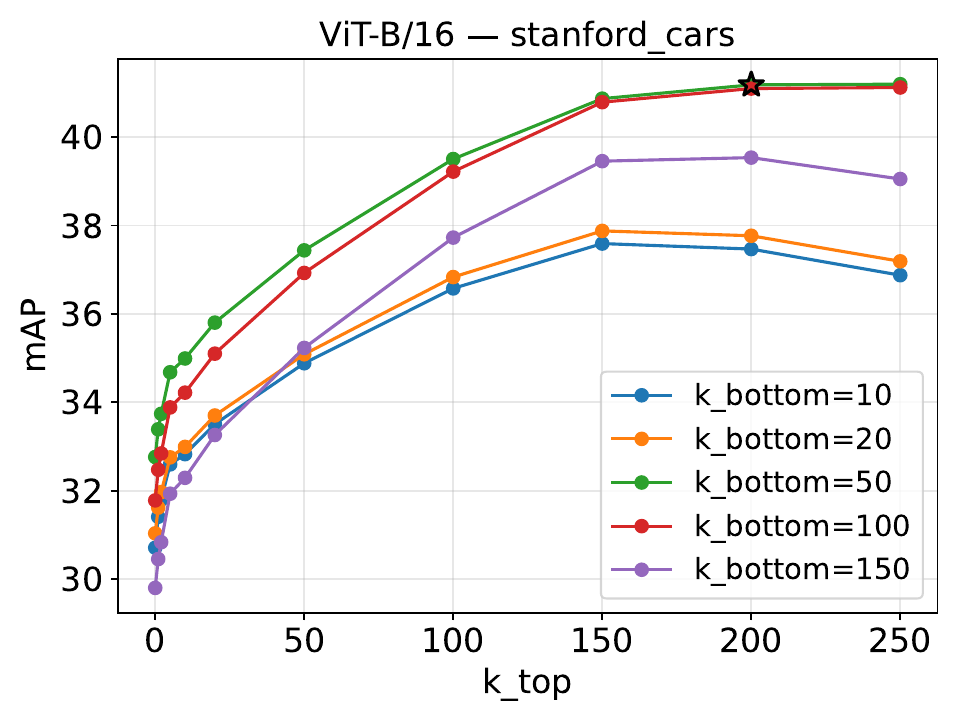}
    \includegraphics[width=0.32\textwidth]{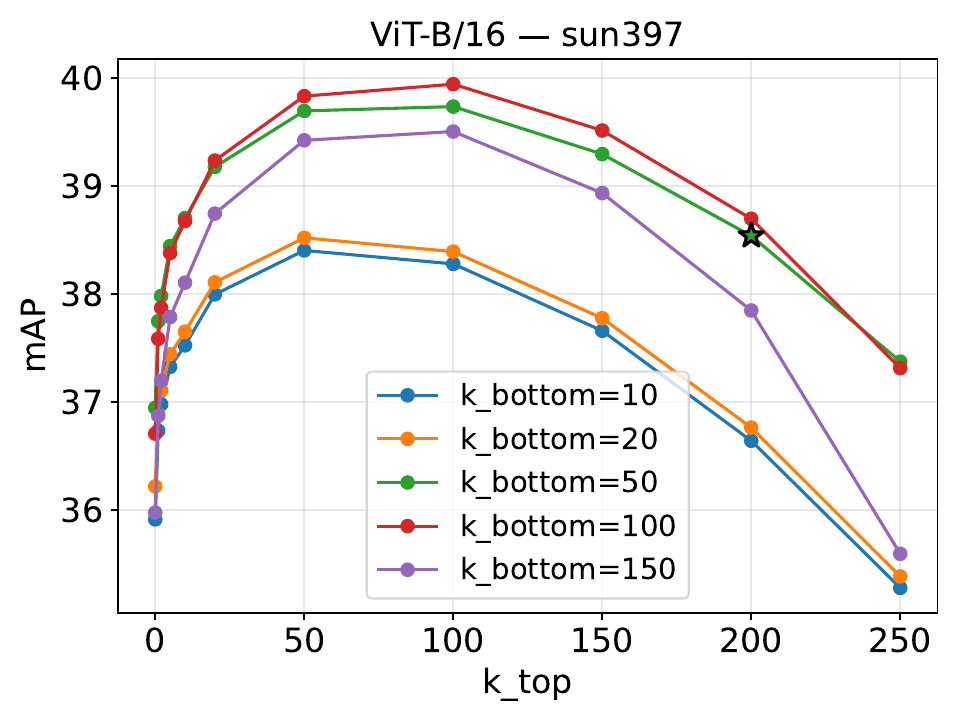}
    \includegraphics[width=0.32\textwidth]{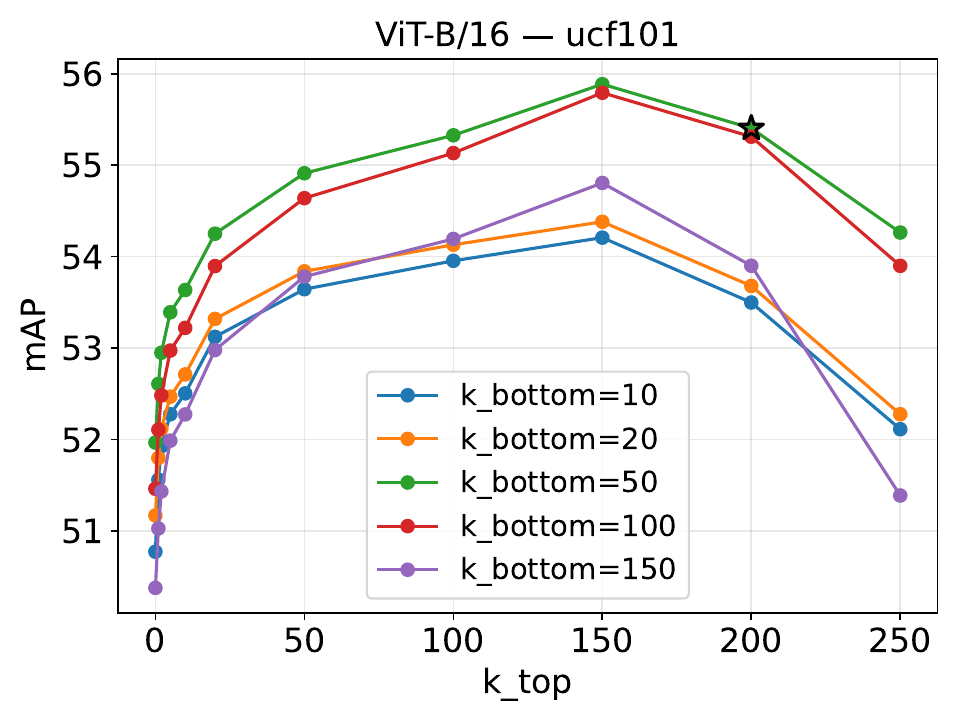}

    \caption{Analysis showing the impact of varying $k_t$ and $k_b$ for the middle band selection across datasets for image-to-image retrieval. The values used in our reported results based on selection from Caltech101 are denoted with stars.}
    \label{fig:supp-hyperparams_image}
\end{figure*}

Figure~\ref{fig:supp-hyperparams_image} shows that, although these values are not  necessarily optimal for every dataset, they generalize well and consistently yield strong performance across all datasets. These plots also indicate that dataset-specific tuning of $(k_t, k_b)$ could further improve performance for some datasets if desired. A similar behavior is observed for text-to-text retrieval in Fig.~\ref{fig:supp-hyperparams_text}, where varying $(k_t, k_b)$ produces similar trends across datasets.

\section{Inter-modal Degradation after Applying IsoCLIP} 
 \label{sec:supp-H}
 
We empirically observe that replacing CLIP projectors with IsoCLIP ones (Eq.~\ref{eq:supp-projector-weights-iso}) reduces CLIP performance on inter-modal tasks such as text-to-image retrieval. This is expected, since the inter-modal operator used to compute inter-modal similarities is explicitly optimized during CLIP training for this purpose, making the original projectors optimal for inter-modal tasks. However, because IsoCLIP modifies only the projector weights, it remains computationally efficient. 

In practical settings where a single image gallery is used for both text–image and image–image retrieval, one can store the pre-projection embeddings in the gallery and use the original CLIP projectors for inter-modal similarity while applying IsoCLIP projectors for intra-modal similarity. This introduces only minimal overhead compared to standard zero-shot CLIP inference, since it requires only an additional matrix multiplication with the projection weights.

\clearpage
\section{The IsoCLIP Algorithm}
In this section we provide the complete pseudocode for applying IsoCLIP to pre-trained CLIP models for intra-modal retrieval tasks. The method consists of a one-time preprocessing step that decomposes and aligns the projector weights with the isotropic subspace, followed by the actual retrieval procedure.
\begin{algorithm}[h]
\caption{IsoCLIP Projector Alignment (Training-Free)}
\label{alg:supp-isoclip-preprocessing}
\begin{algorithmic}[1]
\Require Pre-trained CLIP projectors $W_i \in \mathbb{R}^{d \times d_i}$, $W_t \in \mathbb{R}^{d \times d_t}$
\Require Hyperparameters: $k_t$ (top directions to remove), $k_b$ (bottom directions to remove)
\Ensure Aligned projectors $\widehat{W}_i$, $\widehat{W}_t$
\State \textbf{// Step 1: Construct inter-modal operator}
\State $\Psi \gets W_i^{\top} W_t \in \mathbb{R}^{d_i \times d_t}$
\State
\State \textbf{// Step 2: Singular Value Decomposition}
\State $U, \Sigma, V^{\top} \gets \text{SVD}(\Psi)$ \Comment{$U \in \mathbb{R}^{d_i \times r}$ (image-side), $V \in \mathbb{R}^{d_t \times r}$ (text-side), $\Sigma \in \mathbb{R}^{r \times r}$}
\State
\State \textbf{// Step 3: Select middle-band isotropic subspace}
\State Identify the subspace $\mathcal{S}_U = \text{span}\{u_j \mid j \in [k_t, r - k_b]\}$ 
       \Comment{Image subspace}

\State $U_{\mathcal{S}_U} \gets [u_{k_t}, u_{k_t+1}, \ldots, u_{r-k_b}]$ 
       \Comment{Extract columns from $U$}

\State  

\State Identify the subspace $\mathcal{S}_V = \text{span}\{v_j \mid j \in [k_t, r - k_b]\}$ 
       \Comment{Text subspace}

\State $V_{\mathcal{S}_V} \gets [v_{k_t}, v_{k_t+1}, \ldots, v_{r-k_b}]$ 
       \Comment{Extract columns from $V$}
\State
\State \textbf{// Step 4: Align projectors to the isotropic subspace}
\State $\widehat{W}_i \gets W_i U_{\mathcal{S}_U} U_{\mathcal{S}_U}^{\top}$ \Comment{Project image projector}
\State $\widehat{W}_t \gets W_t V_{\mathcal{S}_V} V_{\mathcal{S}_V}^{\top}$ \Comment{Project text projector}
\State
\State \Return $\widehat{W}_i$, $\widehat{W}_t$
\end{algorithmic}
\end{algorithm}

\begin{algorithm}[h]
\caption{IsoCLIP for Image-to-Image Retrieval }
\label{alg:supp-isoclip-image-efficient}
\begin{algorithmic}[1]
\Require Aligned image projector $\widehat{W}_i$ (from Algorithm~\ref{alg:supp-isoclip-preprocessing})
\Require Image encoder $f_{\theta}(\cdot)$
\Require Query image $i_q$
\Require Projected gallery features $\{\widehat{F}_{i_1}, \widehat{F}_{i_2}, \ldots, \widehat{F}_{i_N}\}$ where $\widehat{F}_{i_n} = \widehat{W}_i f_{\theta}(i_n)$
\Ensure Ranked list of gallery images
\State \textbf{// Step 1: Extract and project query feature}
\State $f_{i_q} \gets f_{\theta}(i_q)$ \Comment{Extract pre-projection query feature $\in \mathbb{R}^{d_i}$}
\State $\widehat{F}_{i_q} \gets \widehat{W}_i f_{i_q}$ \Comment{Project to aligned subspace $\in \mathbb{R}^{d}$}
\State
\State \textbf{// Step 2: Compute cosine similarities with pre-computed gallery}
\For{$n = 1$ to $N$}
    \State $s_n \gets \dfrac{\widehat{F}_{i_q}^{\top} \widehat{F}_{i_n}}{\|\widehat{F}_{i_q}\|_2 \, \|\widehat{F}_{i_n}\|_2}$ \Comment{IsoCLIP similarity}
\EndFor
\State
\State \textbf{// Step 3: Rank by similarity}
\State \Return Gallery images ranked by $\{s_1, s_2, \ldots, s_N\}$ in descending order
\end{algorithmic}
\end{algorithm}

\begin{algorithm}[t]
\caption{IsoCLIP for Text-to-Text Retrieval}
\label{alg:supp-isoclip-text-efficient}
\begin{algorithmic}[1]
\Require Aligned text projector $\widehat{W}_t$ (from Algorithm~\ref{alg:supp-isoclip-preprocessing})
\Require Text encoder $g_{\phi}(\cdot)$
\Require Query text $t_q$
\Require  Projected gallery features $\{\widehat{G}_{t_1}, \widehat{G}_{t_2}, \ldots, \widehat{G}_{t_M}\}$ where $\widehat{G}_{t_m} = \widehat{W}_t g_{\phi}(t_m)$
\Ensure Ranked list of gallery texts
\State \textbf{// Step 1: Extract and project query feature}
\State $g_{t_q} \gets g_{\phi}(t_q)$ \Comment{Extract pre-projection query feature $\in \mathbb{R}^{d_t}$}
\State $\widehat{G}_{t_q} \gets \widehat{W}_t g_{t_q}$ \Comment{Project to aligned subspace $\in \mathbb{R}^{d}$}
\State
\State \textbf{// Step 2: Compute cosine similarities with pre-computed gallery}
\For{$m = 1$ to $M$}
    \State $s_m \gets \dfrac{\widehat{G}_{t_q}^{\top} \widehat{G}_{t_m}}{\|\widehat{G}_{t_q}\|_2 \, \|\widehat{G}_{t_m}\|_2}$ \Comment{IsoCLIP similarity}
\EndFor
\State
\State \textbf{// Step 3: Rank by similarity}
\State \Return Gallery texts ranked by $\{s_1, s_2, \ldots, s_M\}$ in descending order
\end{algorithmic}
\end{algorithm}

\end{document}